\documentclass[10pt,journal,compsoc]{IEEEtran}
\ifCLASSOPTIONcompsoc
  \usepackage[nocompress]{cite}
\else
  \usepackage{cite}
\fi
\usepackage{balance}
\usepackage{amsmath,amssymb,amsfonts}
\usepackage{algorithmic}
\usepackage{algorithm}
\usepackage{array}
\usepackage[caption=false,font=normalsize,labelfont=sf,textfont=sf]{subfig}
\usepackage{textcomp}
\usepackage{stfloats}
\usepackage{url}
\usepackage{verbatim}
\usepackage{multirow}
\usepackage{cite}
\usepackage{graphicx}
\usepackage{picinpar}
\usepackage{url}
\usepackage{flushend}
\usepackage[latin1]{inputenc}
\usepackage{colortbl}
\usepackage{soul}
\usepackage{pifont}
\usepackage{color}
\usepackage{alltt}
\usepackage[colorlinks]{hyperref}
\usepackage{enumerate}
\usepackage{siunitx}
\usepackage{breakurl}
\usepackage{epstopdf}
\usepackage{pbox}
\usepackage{amssymb}
\usepackage{threeparttable}
\usepackage{booktabs}
\usepackage{array}
\usepackage[dvipsnames]{xcolor}

\ifCLASSINFOpdf
\else
\fi
\begin{document}
%
\title{Diff9D: Diffusion-Based Domain-Generalized\\Category-Level 9-DoF Object Pose Estimation}
%
%
%
%

\author{Jian Liu,~\IEEEmembership{Member,~IEEE,} Wei Sun, Hui Yang, Pengchao Deng, Chongpei Liu, \\Nicu Sebe,~\IEEEmembership{Senior Member,~IEEE}, Hossein Rahmani, and Ajmal Mian,~\IEEEmembership{Senior Member,~IEEE}

\IEEEcompsocitemizethanks{\IEEEcompsocthanksitem Jian Liu, Wei Sun, Hui Yang, and Chongpei Liu are with the National Engineering Research Center for Robot Visual Perception and Control Technology, College of Electrical and Information Engineering, School of Robotics, and the State Key Laboratory of Advanced Design and Manufacturing for Vehicle Body, Hunan University, Changsha 410082, China. E-mail: (jianliu, wei\_sun, huiyang, chongpei56)@hnu.edu.cn
\IEEEcompsocthanksitem Pengchao Deng is with the Institute of Artificial Intelligence and Robotics, Xi'an Jiaotong University, Xi'an 710049, China. E-mail: dpc987003425@stu.xjtu.edu.cn
\IEEEcompsocthanksitem Nicu Sebe is with the Department of Information Engineering and Computer Science, University of Trento, Trento 38123, Italy. E-mail: sebe@disi.unitn.it
\IEEEcompsocthanksitem Hossein Rahmani is with the School of Computing and Communications, Lancaster University, LA1 4YW, United Kingdom. E-mail: h.rahmani@lancaster.ac.uk
\IEEEcompsocthanksitem Ajmal Mian is with the Department of Computer Science, The University of Western Australia, WA 6009, Australia. E-mail: ajmal.mian@uwa.edu.au.
}
\thanks{This work was done while Jian Liu and Chongpei Liu were visiting Ph.D. students with The University of Western Australia and the University of Trento, respectively, supervised by Prof. Ajmal Mian and Prof. Nicu Sebe.}}

%
%

\markboth{IEEE TRANSACTIONS ON PATTERN ANALYSIS AND MACHINE INTELLIGENCE}%
{Shell \MakeLowercase{\textit{et al.}}: Bare Demo of IEEEtran.cls for Computer Society Journals}

\IEEEtitleabstractindextext{%
\begin{abstract}
Nine-degrees-of-freedom (9-DoF) object pose and size estimation is crucial for enabling augmented reality and robotic manipulation. Category-level methods have received extensive research attention due to their potential for generalization to intra-class unknown objects. However, these methods require manual collection and labeling of large-scale real-world training data. To address this problem, we introduce a diffusion-based paradigm for domain-generalized category-level 9-DoF object pose estimation. Our motivation is to leverage the latent generalization ability of the diffusion model to address the domain generalization challenge in object pose estimation. This entails training the model exclusively on rendered synthetic data to achieve generalization to real-world scenes. We propose an effective diffusion model to redefine 9-DoF object pose estimation from a generative perspective. Our model does not require any 3D shape priors during training or inference. By employing the Denoising Diffusion Implicit Model, we demonstrate that the reverse diffusion process can be executed in as few as 3 steps, achieving near real-time performance. Finally, we design a robotic grasping system comprising both hardware and software components. Through comprehensive experiments on two benchmark datasets and the real-world robotic system, we show that our method achieves state-of-the-art domain generalization performance. Our code will be made public at \href{https://github.com/CNJianLiu/Diff9D}{https://github.com/CNJianLiu/Diff9D}.
\end{abstract}

}

\maketitle

\IEEEdisplaynontitleabstractindextext

\IEEEpeerreviewmaketitle

\IEEEraisesectionheading{\section{Introduction}\label{Introduction}}
\IEEEPARstart{N}{ine}-degrees-of-freedom (9-DoF) object pose and size estimation predicts the three-dimensional (3D) translation and 3D rotation of an object relative to the camera coordinate system as well as its 3D size. This is a core problem in augmented reality and robotic 3D scene understanding \cite{liu2024survey,1,2,3,4}.

\par Existing pose estimation approaches can be divided into instance-level and category-level methods. Instance-level methods\cite{5,6,7,8,9,10,11,12,13,14,15,16,17} are restricted to specific objects the model has been trained on, which greatly limits their practical applicability. Category-level pose estimation methods exhibit a degree of flexibility, and are able to estimate the pose of novel objects {\em within} categories that are seen during training. Wang \emph{et al.}\cite{18} proposed the first category-level method. Their approach involves the design of a Normalized Object Coordinate Space (NOCS) and the use of the Umeyama algorithm for recovering object pose. However, NOCS exhibits low accuracy due to its inability to effectively represent the diverse shape variations among intra-class objects. 

\par In response to the aforementioned challenge, some shape prior-based methods have been proposed\cite{19,20,21,22,23,24}. Although these methods significantly improve accuracy, they are not trained in an end-to-end fashion. Specifically, these approaches first need to extract the 3D shape prior based on the CAD models of intra-class known objects in offline mode. Then, they estimate the NOCS shape of the intra-class unknown object using shape deformation. Finally, the Umeyama algorithm is used to perform point cloud registration to find the object pose. Constructing CAD model libraries is time-consuming and requires significant manual effort. To address these problems, some prior-free methods have been introduced\cite{27,28,29,30,31,32} to directly regress object pose, achieving better real-time performance during inference. However, these methods still require large amounts of real-world annotated data for training, which is expensive to obtain.

\par In a recent development, to address the challenge of inadequate real-world training data, some domain adaptation methods \cite{35,37,38,39} and a test-time adaptation method \cite{40} have been proposed. The domain adaptation methods require both labeled synthetic data and unlabeled real-world data for training, whereas the test-time adaptation method solely relies on labeled synthetic data during the training process. Nevertheless, the performance of these methods is limited by the huge domain gap between the rendered synthetic domain and the real world.

\begin{figure}[t!]\centering
	\includegraphics[width=\columnwidth]{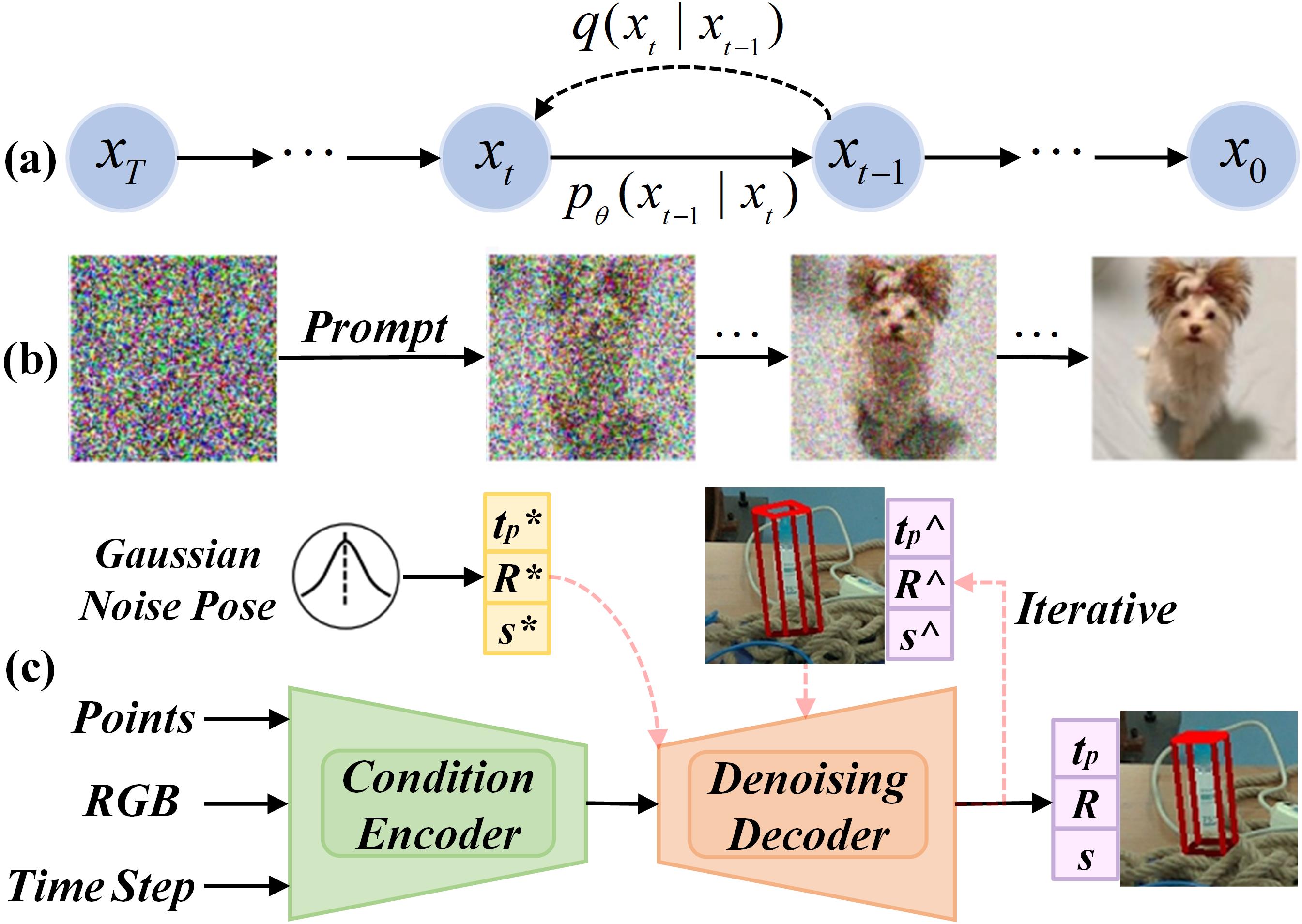}
        \vspace{-2em}
	\caption{{\bf{Comparison of diffusion model-based image generation and object pose estimation}}. {\bf{(a)}}: The process of diffusion model, where ${q}$ and ${p_\theta}$ represent the forward (noising) and reverse (denoising) diffusion processes, respectively. {\bf{(b)}}: Diffusion model-based image generation task, which generates an image based on a prompt. {\bf{(c)}}: The Overall pipeline of our Diff9D, which redefines the 9-DoF object pose and size estimation task from a generative perspective, i.e., from Gaussian noise pose to true object pose. ${t_p}$, $R$, and $s$ represent 3D translation, 3D rotation, and 3D size, respectively.}
\label{Fig1}
\vspace{-1em}
\end{figure}

\par Inspired by Noble Laureate Richard Feynman's quote \emph{``What I cannot create, I do not understand"}, we propose a novel paradigm to redefine object pose estimation from a generative perspective, termed Diff9D. Figure \ref{Fig1} illustrates an overview of the proposed Diff9D. $*$ and $ \wedge $ denote the Gaussian noise pose and an intermediate pose of the denoising process, respectively. Our motivation is to leverage the latent generalization ability of the diffusion model to address the domain generalization challenge \cite{DG} in object pose estimation. Specifically, we introduce a Denoising Diffusion Probabilistic Model (DDPM)-based method for domain-generalized category-level object pose estimation, which is simple yet effective and does not rely on the use of any 3D shape priors during training or inference, facilitating generalization across various object categories. A major challenge in taking a generative modeling approach to pose estimation in robotics is that real-time performance is not feasible since the reverse diffusion requires a large number of denoising steps that must be performed sequentially. We address this challenge by leveraging a Denoising Diffusion Implicit Model (DDIM) \cite{50} and achieving reverse diffusion in as few as 3 steps, enabling near real-time performance. Our main contributions and highlights are as follows:
\vspace{-0.5em}
\begin{itemize} 
\item{We propose a DDPM-based method for domain-generalized category-level 9-DoF object pose and size estimation. Our method redefines the pose estimation problem from a generative perspective to reduce the impact of domain gap. Our model is trained solely on rendered synthetic data and yet generalizes to real-world data, eliminating the laborious human effort required for data collection and annotation.}

\item{We design a simple yet effective object pose/size diffusion model to directly diffuse the sparse pose data, achieving near real-time performance by leveraging the DDIM to perform reverse diffusion in as few as 3 steps. Our model is lightweight and does not require any 3D shape priors during training or inference. Specifically, we perform condition extraction based on the lightweight ResNet18 and PointNet models, then propose a transformer-based denoiser for denoising.}

\item{We build a robotic grasping system and deploy the proposed method on it. Extensive experiments on real-world robotic grasping scenes and two widely used challenging datasets (REAL275 and Wild6D) demonstrate that the proposed method achieves superior domain generalization performance, which is able to generalize to real-world grasping tasks at 17.2 frames per second (FPS).}
\end{itemize}
\vspace{-0.5em}
\par The rest of this paper is organized as follows. The next section reviews related works. Sec. \ref{Methodology} presents the proposed method and Sec. \ref{Designed Robotic Grasping System} presents the designed robotic grasping system, including hardware/software setup and workflow. Next, extensive experimental results are reported in Sec. \ref{Experiments} to demonstrate the superior performance of the proposed method. Finally, Sec. \ref{Conclusion} concludes the paper.

\vspace{-1.25em}
\section{Related Work}\label{Related Work}
This section first reviews the object pose and size estimation methods, dividing them into instance-level and category-level methods, and then reviews recent diffusion model-based methods and explains how our proposed method differs from them. Finally, we review the object pose estimation-based robotic grasping methods.

\vspace{-1.25em}
\subsection{Instance-Level Methods}
Instance-level methods are trained on known objects \cite{SSD-6D} and can be mainly divided into three categories: correspondence-based, template-based, and direct regression-based. Correspondence-based methods can be further divided into 2D-3D correspondence and 3D-3D correspondence. 2D-3D correspondence methods\cite{5,6} first define the keypoints between RGB image and object CAD model. This is followed by training a model to predict the 2D keypoints and using the Perspective-n-Points (PnP) algorithm to solve the object pose. 3D-3D correspondence methods\cite{7,8} define the keypoints on the object CAD model directly and use the observed point cloud to predict the predefined 3D keypoints. Next, they apply the least squares algorithm to solve the object pose. However, most correspondence-based methods rely heavily on rich texture information and may not work well when applied to textureless objects.

\par There are some point cloud-based template methods, which are based on point cloud registration \cite{9,10}. Specifically, the template is the object CAD model with the canonical pose, and the purpose of these methods is to find the optimal relative pose that aligns the observed point cloud with the template. Besides these methods, RGB-based template methods\cite{11,12} also exist, which require collecting and annotating object images from various perspectives during the training phase to create templates. After that, these methods train a template matching model to find the closest template to the observed image and use the template pose as the actual pose of the object. Overall, template-based methods can be effectively applied to textureless objects, however, the template-matching process is generally time-consuming.

\par With the rapid advancement of deep learning technology, direct regression-based methods\cite{13,14,15,16,17} have recently gained popularity. These methods use the ground-truth object poses for supervision and train models to regress the object pose end-to-end. Specifically, DenseFusion\cite{13} fuses the RGB and depth features and proposes a pixel-level dense fusion network for pose regression. FFB6D\cite{14} further designs a bidirectional feature fusion network to fully fuse the RGB and depth features. GDR-Net\cite{15} proposes a geometry-guided network for end-to-end monocular object pose regression. HFF6D\cite{16} designs a hierarchical feature fusion framework for object pose tracking in dynamic scenes. Although instance-level methods have achieved high accuracy, they are restricted to fixed instances, meaning that they only work for specific objects on which they are trained.

\vspace{-1em}
\subsection{Category-Level Methods}
Research in the domain of category-level methods has received substantial attention given their potential for generalization to unknown objects within the given object categories. 
NOCS\cite{18} introduces a normalized object coordinate space, providing a standardized representation for a category of objects, and recovers object pose using the Umeyama algorithm. SPD\cite{19} leverages shape prior deformation to solve the problem of diverse shape variations between intra-class objects. Due to the superior performance achieved by SPD, other prior-based methods are also subsequently proposed. CR-Net\cite{20} designs a recurrent framework for iterative residual refinement to improve the shape prior-based deformation and coarse to fine object pose estimation. SGPA\cite{21} utilizes the structure similarity between the shape prior and the observed intra-class unknown object to dynamically adapt the shape prior. 6D-ViT\cite{22} introduces Pixelformer and Pointformer networks, based on the Transformer architecture, to extract more refined features of the observed objects. STG6D\cite{23} goes a step further and fuses the difference features between the shape prior and the observed objects, enabling more refined deformation. RBP-Pose\cite{24} designs a geometry-guided residual object bounding box projection network to solve the insufficient pose-sensitive feature extraction.
CATRE\cite{26} proposes a pose refinement method based on the alignment of the observed point cloud and the shape prior, which can be used to further refine the object pose estimated by the above methods. GeoReF \cite{25} builds upon CATRE \cite{26} to tackle the geometric variation issue by incorporating hybrid scope layers and learnable affine transformations. Although prior-based methods significantly improve accuracy, constructing CAD model libraries is cumbersome and time-consuming. 

\par Besides these prior-based methods, DualPoseNet\cite{27} introduces a dual pose encoder with refined learning of pose consistency and regresses object pose via two parallel pose decoders. FS-Net\cite{28} proposes a shape-based 3D graph convolution network and performs decoupled regression for translation, rotation, and size. GPV-Pose\cite{29} harnesses geometry-guided point-wise voting to enhance the learning of category-level pose-sensitive features. HS-Pose\cite{30} further proposes a hybrid scope feature extraction network, addressing the limitations associated with the size and translation invariant properties of 3D graph convolution. IST-Net\cite{31} explores the necessity of shape priors for category-level pose estimation and proposes an implicit space transformation-based prior-free method. VI-Net\cite{32} addresses the problem of poor rotation estimation by decoupling rotation into viewpoint and in-plane rotations. While these methods do not depend on shape priors, they still require large amounts of real-world annotated data for training, which hinders their practical applicability.

\par To address the problem of insufficient real-world training data, CPPF\cite{33} performs pose estimation in the wild by introducing a category-level point pair feature voting method. SAR-Net\cite{34} proposes to explore the shape alignment of each intra-class unknown object against its corresponding shape prior without using real-world training data. SSC6D\cite{35} proposes a self-supervised method using DeepSDF\cite{36} for deep implicit shape representation. UDA-COPE\cite{37} utilizes a teacher-student self-supervised learning framework to achieve domain adaptation. RePoNet\cite{38} proposes a self-supervised method based on pose and shape differentiable rendering. DPDN\cite{39} designs a parallel deep prior deformation-based domain generalization learning scheme. More recently, TTA-COPE\cite{40} introduces a test-time adaptation method, which initially trains the model on labeled synthetic data and subsequently utilizes the pretrained model for test-time adaptation in real-world data during inference. Nevertheless, the performance of these methods is limited by the huge domain gap between the rendered synthetic domain and the real world.

\begin{figure*}[htbp]
\centering
\includegraphics[width=\textwidth]{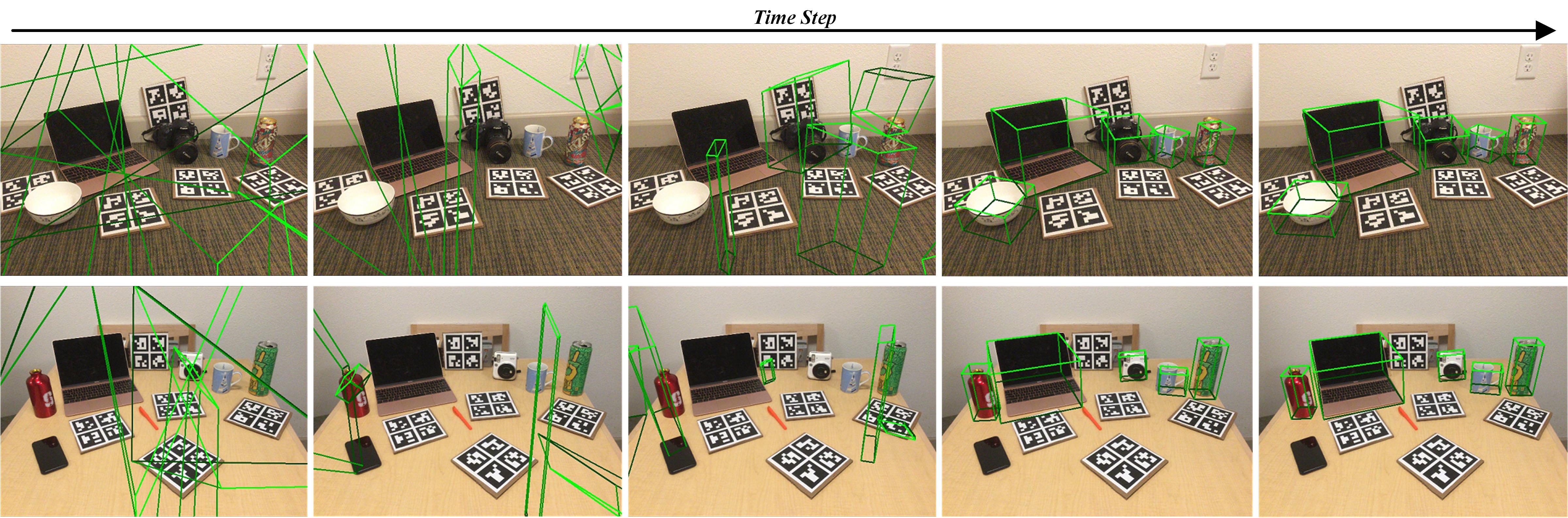}
\vspace{-2em}
\caption{Some visualizations of the reverse diffusion process, representing the diffusion from Gaussian noise poses to objects poses in the observed scene.}
\label{Fig2}
\vspace{-1em}
\end{figure*}

\vspace{-1em}
\subsection{Diffusion Model-Based Methods}
\par More recently, diffusion models gained popularity in object pose estimation. In terms of instance-level methods, DiffusionReg \cite{DiffusionReg} proposes a point cloud registration framework leveraging the SE(3) diffusion model. This model gradually perturbs the optimal rigid transformation of a pair of point clouds by continuously injecting perturbations through the SE(3) forward diffusion process. The SE(3) reverse denoising process is then used to progressively denoise, approaching the optimal transformation for precise pose estimation. 6D-Diff \cite{6D-Diff} develops a diffusion-based framework that formulates 2D keypoint detection as a denoising process, enabling more accurate 2D-3D correspondences. As for category-level methods, GenPose \cite{42} introduces a score-based diffusion model to tackle the multi-hypothesis issue in symmetric objects and partial point clouds. Their approach first uses the score-based diffusion model to generate multiple pose candidates and then employs an energy-based diffusion model to eliminate abnormal poses. DiffusionNOCS \cite{DiffusionNOCS} first diffuses the NOCS map of the object using multi-modal input as a condition, and then uses an offline registration algorithm to align and solve the object pose.

\par In general, DiffusionReg \cite{DiffusionReg} and 6D-Diff \cite{6D-Diff} are instance-level methods. GenPose \cite{42} and DiffusionNOCS \cite{DiffusionNOCS} mainly focus on 6-DoF pose (excluding 3D size). Moreover, GenPose does not focus on solving the problem of domain generalization, and the diffusion target of DiffusionNOCS is the NOCS map. Different from the above methods, we aim to develop a category-level 9-DoF object pose estimation method suitable for real-world robotic applications using only rendered synthetic data for training. This approach faces two main challenges: \textbf{1)} The significant domain gap between synthetic and real-world data, which adversely affects the performance of conventional regression models. We propose using a denoising diffusion probabilistic model to frame object pose estimation as a generative process. The diffusion model performs extensive sampling on the Markov chain, which can effectively expand the distribution of the synthetic pose data, making the data distribution more uniform \cite{Ed-sam}, thus contributing to reducing the impact of the domain gap on the pose estimation model. \textbf{2)} Efficient pose estimation is crucial due to the limited computational resources available in robotics. To address this problem, we design a simple yet effective network structure, employing lightweight baseline networks (ResNet18 \cite{52} for RGB image and PointNet \cite{53} for point cloud) and using only global features as conditions. Additionally, given the sparsity of object pose data (only 15 values), our approach differs from dense diffusion tasks like image generation and can be efficient with fewer diffusion steps.

\vspace{-1em}
\subsection{Object Pose Estimation-Based Robotic Grasping}
To investigate the application of object pose estimation technology for robotic grasping, Zhang \emph{et al.}\cite{41} developed a practical robotic grasping system based on pose estimation with protective correction. GenPose \cite{42} proposes a score-based diffusion method for 6-DoF object pose estimation and explores its application for robotic manipulation. Liu \emph{et al.}\cite{43} introduced a difference-aware shape adjustment method based on fine segmentation. They also built a robotic grasping platform to verify the practical performance of the pose estimation. For applications where depth images are not practical, e.g., under strong or low light conditions or for transparent and reflective objects, Wolnitza \emph{et al.}\cite{44} proposed a monocular method for 3D object reconstruction and object pose estimation and used it for robotic grasping. BDR6D\cite{45} is another method that first predicts the depth information from a monocular image and then utilizes a bidirectional depth residual network for pose estimation. The proposed method is then deployed with a UR5 robot to perform grasping and manipulating tasks. More recently, STG6D\cite{23} develops a robotic continuous grasping system via a category-level method and proposes a pre-defined vector orientation-based grasping strategy. DGPF6D\cite{46} introduces a contrastive learning-guided shape prior-free category-level method for domain-generalized robotic picking. Yu \emph{et al.}\cite{47} proposed a self-supervised-based category-level object pose estimation method for robotic grasping. Chen \emph{et al.}\cite{48} explored a sim-to-real method by iterative self-training for robotic bin picking.

\par Side-stepping from the above object pose estimation methods for robotic grasping, this paper proposes a DDPM-based novel paradigm for domain-generalized category-level 9-DoF object pose estimation, redefining the pose estimation process from a generative perspective. Leveraging the latent generalization ability of the diffusion model, the proposed method achieves training solely with rendered synthetic images for generalization to real-world robotic grasping scenarios.

\vspace{-0.5em}
\section{Methodology}\label{Methodology}
This section gives details of the proposed Diff9D. First, we illustrate the pose diffusion process (Sec. \ref{Pose Diffusion}), and then describe the condition extraction for pose diffusion (Sec. \ref{Condition Extraction for Pose Diffusion}). Next, we introduce the proposed transformer-based denoiser for pose denoising (Sec. \ref{Self-Attention U-Net for Pose Denoising}). Finally, we elaborate the supervision method (Sec. \ref{Loss Function}).

\vspace{-1em}
\subsection{Pose Diffusion}\label{Pose Diffusion}
\subsubsection{Forward Pose Diffusion Process}
Given a 9-DoF object pose sampled from a real-world pose distribution ${x_0} \sim q\left( x \right)$, we define a forward diffusion process where Gaussian noise is gradually added to the sample in $T$ time steps ($T$ represents the maximum time step), producing a sequence of noisy samples ${x_1}, \cdot  \cdot  \cdot ,{x_T}$. The time step is controlled by a variance schedule $\left\{ {{\beta _t} \in \left( {0,1} \right)} \right\}_{t = 1}^T$ and ${\beta _1} < {\beta _2} <  \cdot  \cdot  \cdot  < {\beta _T}$. The forward pose diffusion process from ${x_{t - 1}}$ to ${x_t}$ is defined as \cite{49}:
\begin{equation}\label{equation4}
q\left( {{x_t}|{x_{t - 1}}} \right) = {\cal N}\left( {{x_t};\sqrt {1 - {\beta _t}} {x_{t - 1}},{\beta _t}{\rm \emph{I}}} \right),
\end{equation}
where ${\cal N}\left( {\mu ,{\sigma ^2}} \right)$ represents a Gaussian distribution. The step-by-step diffusion process follows the Markov chain assumption:
\begin{equation}\label{equation5}
q\left( {{x_{1:T}}|{x_0}} \right) = \prod\limits_{t = 1}^T {q\left( {{x_t}|{x_{t - 1}}} \right)}.
\end{equation}
Specifically, ${x_t}$ can be represented as:
\begin{equation}\label{equation6}
{x_t} = \sqrt {{\beta _t}} {\varepsilon _t} + \sqrt {1 - {\beta _t}} {x_{t - 1}},
\end{equation}
where ${\varepsilon _t}$ is a randomly sampled standard Gaussian noise at time step $t$. Let ${\alpha _t} = 1 - {\beta _t}$ and ${{\bar \alpha }_t} = \prod\nolimits_{i = 1}^t {{\alpha _i}}$, we can get:
\begin{equation}\label{equation10}
{x_t} = \sqrt {{{\bar \alpha }_t}} {x_0} + \sqrt {1 - {{\bar \alpha }_t}} \varepsilon.
\end{equation}
Hence, the forward pose diffusion process from ${x_{0}}$ to ${x_t}$ can be represented as:
\begin{equation}\label{equation11}
q\left( {{x_t}|{x_0}} \right) = {\cal N}\left( {{x_t};\sqrt {{{\bar \alpha }_t}} {x_0},\left( {1 - {{\bar \alpha }_t}} \right)I} \right).
\end{equation}

\begin{figure*}[htbp]
\centering
\includegraphics[width=\textwidth]{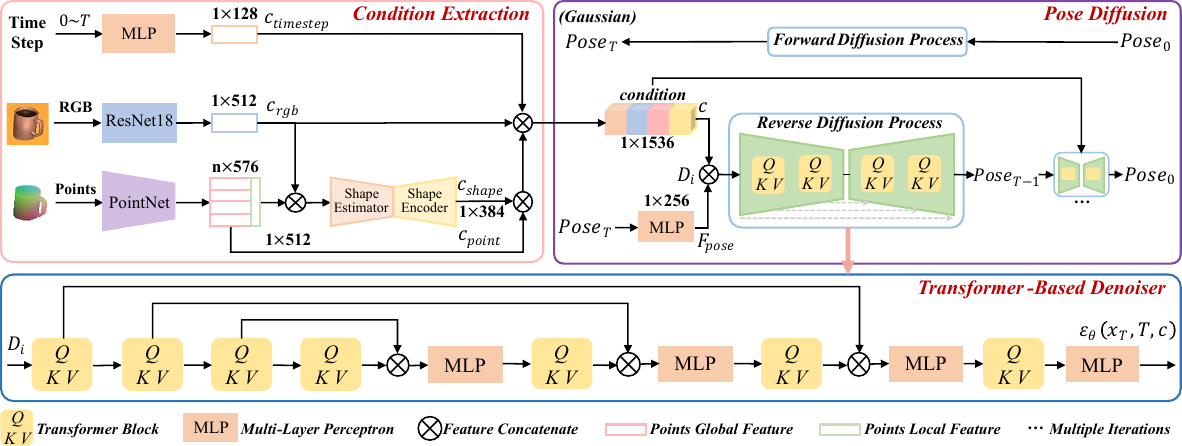}
\vspace{-2em}
\caption{Workflow of the proposed Diff9D, which includes three main parts (pose diffusion, condition extraction for pose diffusion, and transformer-based denoiser for pose denoising). The input of Diff9D is RGB image, point cloud, and time step $T$ and its corresponding noise pose $Pos{e_T}$. Note that the image is first instance segmented by Mask R-CNN \cite{57} before condition extraction. The condition extraction extracts the input condition $c$. The pose diffusion consists of forward (noising) and reverse (denoising) diffusion processes. Forward diffusion continuously adds noise to the ground-truth object pose $Pos{e_0}$. Reverse diffusion first concatenates the noise pose features ${F_{pose}}$ and $c$ to form the input ${D_{i}}$ for the denoiser. The transformer-based denoiser then takes ${D_{i}}$ as input and predicts the pose noise ${\varepsilon _\theta }\left( {{x_T},T,c} \right)$. Finally, ${\varepsilon _\theta }\left( {{x_T},T,c} \right)$ can be used to denoise $Pos{e_T}$ through the reverse diffusion process based on the Markov chain to obtain $Pos{e_{T - 1}}$. We directly use the translation, size, and rotation matrices to represent the object pose, as shown in Fig. \ref{Fig1}. Detailed architecture of the shape estimator and shape encoder is shown in Fig.~\ref{Fig4}.}
\label{Fig3}
\vspace{-1em}
\end{figure*}

\vspace{-1em}
\subsubsection{Reverse Pose Diffusion Process}
As shown in Fig. \ref{Fig3}, the reverse diffusion process aims to recover the object pose from a standard Gaussian noise input ${x_T} \sim {\cal N}\left( {0,I} \right)$. However, obtaining $q\left( {{x_{t - 1}}|{x_t}} \right)$ is not easy, so we learn a model ${p_\theta }$ to approximate this conditional probability to run the reverse diffusion process as:
\begin{equation}\label{equation12}
\begin{array}{l}
{p_\theta }\left( {{x_{0:T}}} \right) = p\left( {{x_T}} \right)\prod\limits_{t = 1}^T {{p_\theta }\left( {{x_{t - 1}}|{x_t}} \right)} ,\\
{p_\theta }\left( {{x_{t - 1}}|{x_t}} \right) = {\cal N}\left( {{x_{t - 1}};{\mu _\theta }\left( {{x_t},t,c} \right),\sum\nolimits_\theta  {\left( {{x_t},t} \right)} } \right),
\end{array}
\end{equation}
where $c$ denotes the condition (see Sec. \ref{Condition Extraction for Pose Diffusion} for more details). Also let ${\alpha _t} = 1 - {\beta _t}$ and ${{\bar \alpha }_t} = \prod\nolimits_{i = 1}^t {{\alpha _i}}$ and follow DDPM\cite{49} to use Bayes' theorem transform  Eq. (\ref{equation12}), then the variance and mean  of ${p_\theta }\left( {{x_{t - 1}}|{x_t}} \right)$ can be parameterized as follows:
\begin{equation}\label{equation13}
\sum\nolimits_\theta  {\left( {{x_t},t} \right)}  = 1/\left( {\frac{{{\alpha _t}}}{{{\beta _t}}} + \frac{1}{{1 - {{\bar \alpha }_{t - 1}}}}} \right) \cdot I = \frac{{1 - {{\bar \alpha }_{t - 1}}}}{{1 - {{\bar \alpha }_t}}} \cdot {\beta _t} \cdot I,
\end{equation}
\begin{equation}\label{equation14}
{\mu _\theta }\left( {{x_t},t,c} \right) = \frac{{\sqrt {{\alpha _t}} \left( {1 - {{\bar \alpha }_{t - 1}}} \right)}}{{1 - {{\bar \alpha }_t}}}{x_t} + \frac{{\sqrt {{{\bar \alpha }_{t - 1}}} {\beta _t}}}{{1 - {{\bar \alpha }_t}}}{x_0}.
\end{equation}
From the previous forward diffusion process Eq. (\ref{equation10}), we can obtain:
\begin{equation}\label{equation16}
{\mu _\theta }\left( {{x_t},t,c} \right) = \frac{1}{{\sqrt {{\alpha _t}} }}\left( {{x_t} - \frac{{1 - {\alpha _t}}}{{\sqrt {1 - {{\bar \alpha }_t}} }}{\varepsilon _\theta }\left( {{x_t},t,c} \right)} \right).
\end{equation}
where ${\varepsilon _\theta }$ denotes the predicted pose noise during the reverse diffusion process.

\begin{figure}[t!]\centering
\includegraphics[width=\columnwidth]{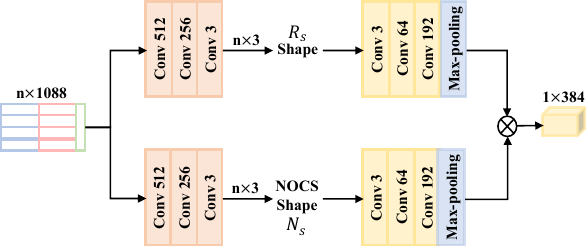}
\vspace{-1.5em}
\caption{Detailed architecture of the shape estimator and shape encoder. We use two parallel branches to estimate and encode shape and NOCS shape.}
\label{Fig4}
\vspace{-1em}
\end{figure}

\par To improve the speed of the reverse diffusion process, we utilize the DDIM\cite{50} scheduler. Some visualizations of the reverse diffusion process are shown in Fig. \ref{Fig2}. Specifically, we take a sample every $T/S$ time steps to reduce the number of sampling time steps from $T$ to $S$. The new sampling schedule is $\left\{ {{\tau _1}, \cdot  \cdot  \cdot ,{\tau _S}} \right\}$. Now, the reverse diffusion process can be expressed as:
\begin{equation}\label{equation18}
\begin{array}{l}
{p_\theta }\left( {{x_{{\tau _{i - 1}}}}|{x_{{\tau _i}}}} \right) = \\
{\cal N}\left( {{x_{{\tau _{i - 1}}}};\sqrt {{{\bar \alpha }_{{\tau _{i - 1}}}}} {x_0} + \sqrt {1 - {{\bar \alpha }_{{\tau _{i - 1}}}} - \sigma _{{\tau _i}}^2} \frac{{{x_{{\tau _i}}} - \sqrt {{{\bar \alpha }_{{\tau _i}}}} {x_0}}}{{\sqrt {1 - {{\bar \alpha }_{{\tau _i}}}} }},\sigma _{{\tau _i}}^2I} \right)
\end{array},
\end{equation}
where $\sigma _t^2$ can be obtained from Eq. (\ref{equation12}) and Eq. (\ref{equation13}) as:
\begin{equation}\label{equation19}
\sigma _{{\tau _i}}^2 = \frac{{1 - {{\bar \alpha }_{{\tau _{i - 1}}}}}}{{1 - {{\bar \alpha }_{{\tau _i}}}}} \cdot {\beta _{{\tau _i}}}.
\end{equation}
\par Overall, the reverse diffusion process predicts the pose noise ${\varepsilon _\theta }\left( {{x_t},t,c} \right)$ by learning a model, and then utilizes the DDIM\cite{50} scheduler for denoising.

\subsection{Condition Extraction for Pose Diffusion}\label{Condition Extraction for Pose Diffusion}
However, it is difficult to directly perform the reverse diffusion process using only ${x_T} \sim {\cal N}\left( {0,I} \right)$ as the input of the diffusion model. Therefore, we propose to incorporate conditional information from the input to guide the diffusion model to achieve more accurate pose predictions. Given that obtaining RGB images and point clouds is straightforward with an RGB-D camera, we utilize them along with the time step as inputs for condition extraction. The detailed architecture is shown in Fig. \ref{Fig3}. Specifically, the diffusion process is associated with the time step. To extract the time step condition $c_{timestep}$, we follow DiffPose\cite{51} to employ a Multi-Layer Perceptron (MLP). Additionally, for the observed RGB image and point cloud, we utilize the lightweight ResNet \cite{52} and PointNet \cite{53} to extract the RGB global condition $c_{rgb}$ and point cloud condition (including global features $c_{point}$ and local features) of the observed object, respectively.

\begin{figure*}[htbp]
\centering
\includegraphics[width=\textwidth]{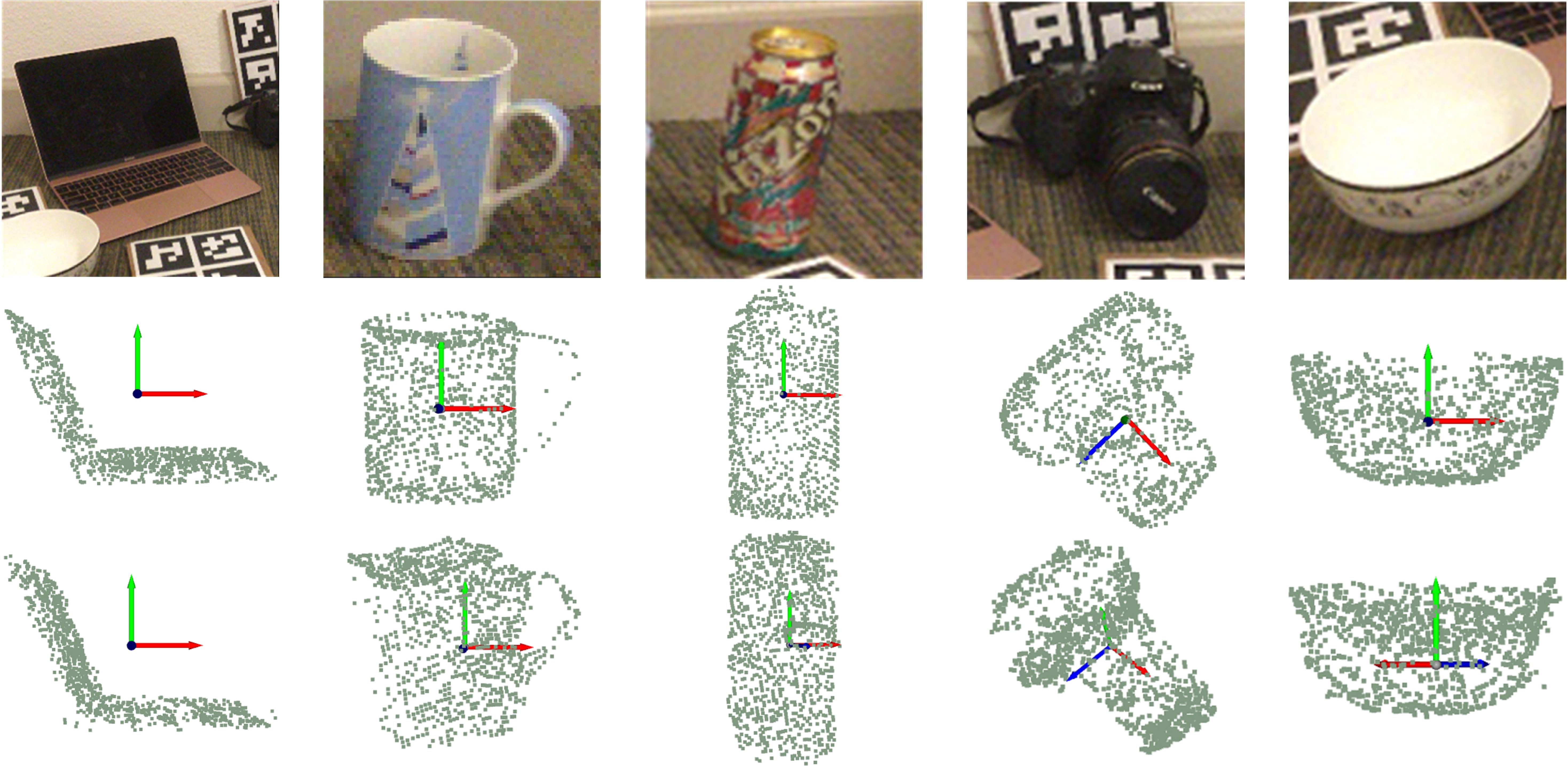}
\vspace{-2em}
\caption{Some visualizations of the estimated shape and NOCS shape. Top, middle, and bottom rows denote the observed RGB images and their corresponding estimated shape and NOCS shape, respectively. It can be seen that these two processes can introduce potential 3D geometric information, making the pose diffusion process geometrically guided.}
\label{Fig5}
\vspace{-1em}
\end{figure*}

\par Inspired by the positive impact of the 3D shape reconstruction on object pose estimation \cite{19,21,23,39}, we point-wise concatenate the point cloud condition with the RGB global condition. The concatenated features are subsequently fed into a shape estimator-encoder network, thereby incorporating supervision for the 3D shape of intra-class unknown objects.
The detailed architecture of the shape estimator-encoder network is shown in Fig. \ref{Fig4}. Specifically, we use two parallel branches to perform decoupled estimation and encoding for the shape and NOCS shape of intra-class unknown objects. Some visualizations are shown in Fig. \ref{Fig5}. Subsequently, we perform max pooling on the encoded two-branch features and concatenate them as the shape condition $c_{shape}$. Finally, we concatenate the obtained time step, RGB global, point cloud global, and shape conditions to obtain the conditional input of pose diffusion as follows:
\begin{equation}\label{equation1}
c = cat\left( {{c_{time step}},{c_{rgb}},{c_{point}},{c_{shape}}} \right).
\end{equation}

\par To supervise the condition extraction process, we follow\cite{19,21,23,39} and use the Chamfer distance between the ground-truth 3D model ${{M_{gt}}}$ and the estimated shape ${R_s}$, and the Smooth-L1 distance between the ground-truth NOCS shape ${{M_{{N_s}}}}$ and the estimated NOCS shape ${N_s}$ as the loss functions. ${{M_{{N_s}}}}$ can be easily obtained from ${{M_{gt}}}$ \cite{18}. Chamfer distance ${L_{cd}}$ can be expressed as:
\begin{equation}\label{equation2}
\begin{array}{l}
{L_{cd}}\left( {{R_s},{M_{gt}}} \right) = \frac{1}{{2n}} \times \\
\left( {\sum\limits_{a \in {R_s}} {\mathop {\min }\limits_{b \in {M_{gt}}} } \parallel a - b\parallel _2^2 + \sum\limits_{b \in {M_{gt}}} {\mathop {\min }\limits_{a \in {R_s}} } \parallel a - b\parallel _2^2} \right),
\end{array}
\end{equation}
where $n$ denotes the number of points. The Smooth-L1 distance ${L_{_{S - L1}}}$ can be expressed as:
\begin{equation}\label{equation3}
\begin{array}{l}
{L_{_{S - L1}}}\left( {{N_s},{M_{{N_s}}}} \right) = \frac{1}{n}\sum\limits_{i = 1}^n {\sum\limits_{k = 1}^3 {\left\{ {\begin{array}{*{20}{c}}
{5{x^2},\begin{array}{*{20}{c}}
{}
\end{array}if\begin{array}{*{20}{c}}
{}
\end{array}x \le 0.1,}\\
{x - 0.05,\begin{array}{*{20}{c}}
{}
\end{array}otherwise,}
\end{array}} \right.} } \\
{\rm{and}}\begin{array}{*{20}{c}}
{}
\end{array}x = \left| {N_s^{ik} - M_{{N_s}}^{ik}} \right|,
\end{array}
\end{equation}
where $k$ denotes the dimension of the coordinates.

\vspace{-1em}
\subsection{Transformer-Based Denoiser for Pose Denoising}\label{Self-Attention U-Net for Pose Denoising}
Since ${{\varepsilon _t}}$ is available at training time, we need to train a network to predict the pose noise conditioned on $c$, i.e., predict ${{\varepsilon _\theta }\left( {{x_t},t,c} \right)}$. Then, the denoising loss term can be parameterized to minimize the difference from ${{\varepsilon _t}}$ to ${{\varepsilon _\theta }\left( {{x_t},t,c} \right)}$ as follows:
\begin{equation}\label{equation20}
\begin{array}{l}
\begin{array}{l}
{L_{diff}} = {\mathbb{E}_{t \sim \left[ {1,T} \right],{x_0},{\varepsilon _t}}}\left[ {{{\left\| {{\varepsilon _t} - {\varepsilon _\theta }\left( {{x_t},t,c} \right)} \right\|}^2}} \right]\\
 = {\mathbb{E}_{t \sim \left[ {1,T} \right],{x_0},{\varepsilon _t}}}\left[ {{{\left\| {{\varepsilon _t} - {\varepsilon _\theta }\left( {\sqrt {{{\bar \alpha }_t}} {x_0} + \sqrt {1 - {{\bar \alpha }_t}} \varepsilon ,t,c} \right)} \right\|}^2}} \right].
\end{array}
\end{array}
\end{equation}

\begin{figure}[t!]\centering
	\includegraphics[width=0.8\columnwidth]{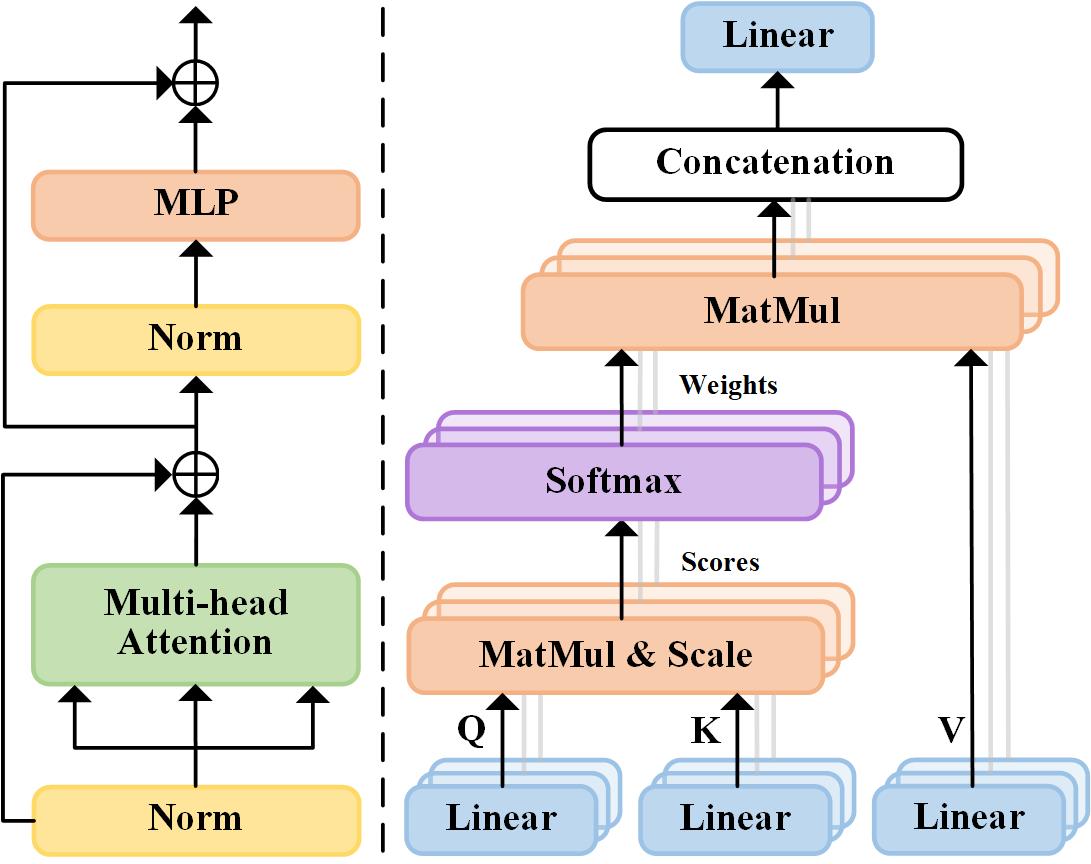}
        \vspace{-0.5em}
	\caption{Detailed architecture of the Transformer block. Left: Transformer block consists of multi-head attention and MLP. $ \oplus $ represents residual connection. Right: Detailed architecture of the multi-head attention mechanism.}
\label{Fig6}
\vspace{-1em}
\end{figure}

\par Due to the sparsity of pose data, utilizing the cross-attention mechanism, commonly employed in other Diffusion models, is not feasible. Consequently, we propose a transformer-based denoiser as the denoising model consisting of multiple self-attention Transformer blocks, MLPs, and skip connections. The detailed architecture is shown in Fig. \ref{Fig3}. Transformer blocks are used to perform self-attention on the input so that each feature dimension in $D_{i}$ interacts with features from all other dimensions through attention mechanisms. The MLPs are used to reduce the dimension of the concatenated features so that the input feature dimension of the next Transformer block remains unchanged. Through skip connections, our model retains more spatial information during the diffusion process. Specifically, the input of the transformer-based denoiser is: 
\begin{equation}\label{equation21}
{D_{i}} = cat{\left( {c,{F_{pose}}} \right)^ \top},
\end{equation}
where ${{F_{pose}}}$ denotes the pose feature, $cat\left( {} \right)$ represents feature concatenation, and $\top$ represents matrix transpose operation.

\par The detailed architecture of the Transformer block is shown in Fig. \ref{Fig6}. Its core part is the multi-head attention. Specifically, we can get the inputs query $Q$, key $K$, and value $V$ as follows:
\begin{equation}\label{equation22}
\begin{array}{l}
{Q^{\left( m \right)}} = FC_1^{\left( m \right)}\left( {LN\left( {D_{i}} \right)} \right),\\
{K^{\left( m \right)}} = FC_2^{\left( m \right)}\left( {LN\left( {D_{i}} \right)} \right),\\
{V^{\left( m \right)}} = FC_3^{\left( m \right)}\left( {LN\left( {D_{i}} \right)} \right),
\end{array}
\end{equation}
where $m$ denotes the number of heads, $LN\left( {} \right)$ represents layer normalization, and $FC\left( {} \right)$ represents fully connected layer. Then, the multi-head self-attention mechanism proceeds as follows:
\begin{equation}\label{equation23}
SA\left( {D_{i}} \right) = F{C_{cat}}\left( {Softmax\left( {{Q^{\left( m \right)}}{K^{{{\left( m \right)}^ \top}}}/\sqrt d } \right){V^{\left( m \right)}}} \right),
\end{equation}
where $d$ denotes the feature dimension of each head, and $F{C_{cat}}\left( {} \right)$ represents the concatenation of the features of all heads, fed to a fully connected layer.

\begin{figure}[t!]\centering
	\includegraphics[width=\columnwidth]{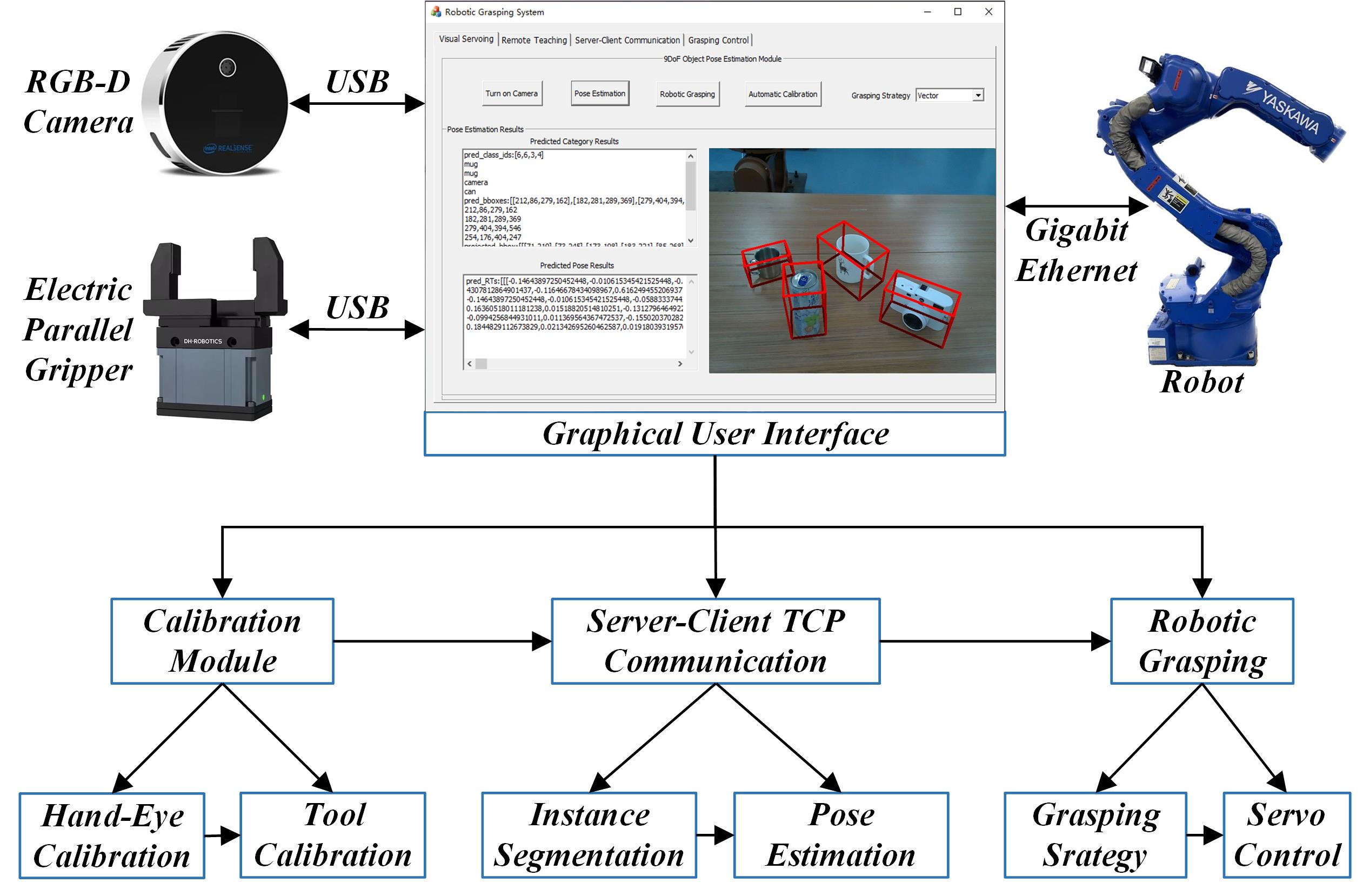}
        \vspace{-2em}
	\caption{Overall setup and workflow of the robotic grasping system. We develop a GUI comprising three parts: calibration module, server-client TCP communication module, and robotic grasping module.}
\label{Fig7}
\end{figure}

\begin{figure}[t!]\centering
\vspace{-1em}
	\includegraphics[width=\columnwidth]{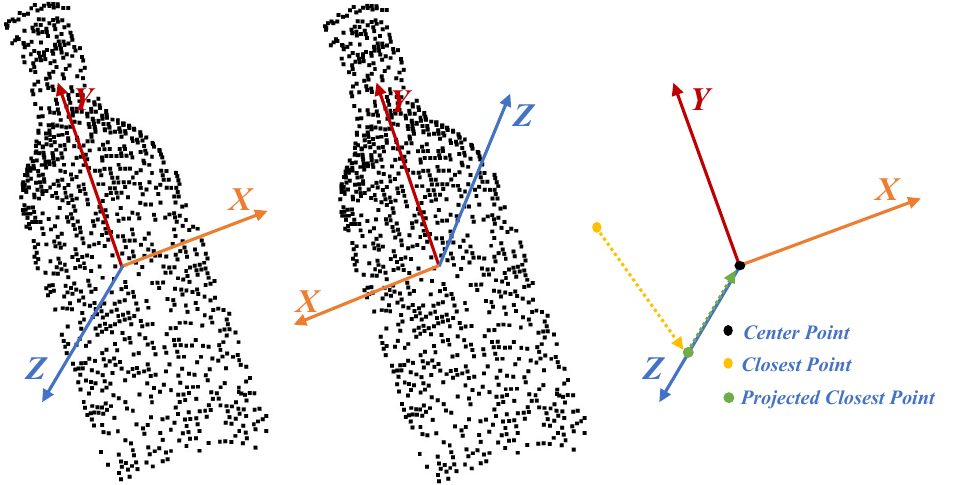}
        \vspace{-2em}
	\caption{Illustration of the robotic grasping strategy. Left and Middle: Ambiguity representation of $Z$-axis and $X$-axis of symmetrical objects. Right: Grasping strategy based on predefined object coordinate system and vector direction.}
\label{Fig8}
\vspace{-1em}
\end{figure}

\par By combining the multi-head self-attention mechanism, MLP, and residual connections, our model is able to better find the global correlation within the features. This makes the diffusion process more robust.

\vspace{-1em}
\subsection{Loss Function}\label{Loss Function}
To perform end-to-end learning, we simply combine shape reconstruction loss, NOCS shape reconstruction loss, and pose diffusion loss as follows:
\begin{equation}\label{equation24}
{L_{total}} = {L_{cd}} + {L_{S - L1}} + {L_{diff}},
\end{equation}
where ${L_{total}}$ is the final loss we choose to optimize the learning of the proposed Diff9D. We directly use the translation, size, and rotation matrices to represent the object pose.

\vspace{-1em}
\section{Robotic Grasping System Design}\label{Designed Robotic Grasping System}
\subsection{Hardware and Software Setup}
\subsubsection{Hardware Composition}
The hardware for the robotic grasping system is composed of an Intel RealSense L515 RGB-D camera, a Yaskawa robot
MOTOMAN-MH12, an electric parallel gripper DH-PGI-140-80, and a host computer. Their mutual connections are illustrated in Fig. \ref{Fig7}. Specifically, the RGB-D camera and electric gripper are connected to the host computer through USB, and the robot is connected to the host computer through Gigabit Ethernet. The installation method between the camera and robot is eye-in-hand.

\vspace{-1em}
\subsubsection{Software Setup}
The configuration of the software for the designed robotic grasping system is shown in Fig. \ref{Fig7}. Since the robotic grasping system consists of three modules (calibration module, server-client TCP communication module, and robotic grasping module), we develop a GUI to connect these three functional modules. The calibration module includes robotic hand-eye calibration and tool calibration. We use the HALCON calibration method and the five-point calibration method to complete hand-eye and tool calibration, respectively. The server-client TCP communication module is used for real-time communication between the server and the client. We set the host computer as the server and set the trained instance segmentation model and pose diffusion model as two clients. Finally, the robotic grasping module includes two parts, i.e., grasping strategy and servo control.

\vspace{-1em}
\subsection{Robotic Grasping Workflow}
\subsubsection{Overall Workflow}
In general, the robotic grasping of objects in 3D space can be completed as follows:
\vspace{-0.5em}
\begin{equation}\label{equation}
{M_{o2t}} = {M_{e2t}}{M_{c2e}}{M_{o2c}},
\vspace{-0.5em}
\end{equation}
where ${M_{o2t}}$ denotes the transformation matrix from the object to the tool coordinate system. This is what the robot needs to obtain for the grasping task. ${M_{e2t}}$ represents the transformation matrix from the robot end coordinate system to the tool coordinate system, which can be solved through tool calibration. ${M_{c2e}}$ denotes the transformation matrix from the camera coordinate system to the robot end coordinate system, which can be solved through hand-eye calibration. ${M_{o2c}}$ denotes the transformation matrix from the object coordinate system to the camera coordinate system, which is \emph{the most significant part} and can be obtained through object pose estimation.

\vspace{-1em}
\subsubsection{Grasping Strategy}
After estimating the poses of all intra-class unknown objects in the scene, we utilize a simple yet effective grasping strategy to achieve continuous grasping of multiple objects. Specifically, we use the depth of the observed object center point (i.e., the average coordinate point) to determine the grasping sequence. Then, the robot grasps it according to the predefined object coordinate system and vector direction. The illustration of the robotic grasping strategy is shown in Fig. \ref{Fig8}. We project the closest point of the observed object onto the $Z$-axis of the predefined object coordinate system, and then connect it to the object center point to form a predefined vector. We select the direction of the predefined vector as the grasping direction, and make the closing direction of the gripper parallel to the $X$-axis. We found that using this simple grasping strategy can simultaneously grasp symmetrical and asymmetrical objects (it is well known the $X$ and $Z$ axes of symmetrical objects have uncertainty, as shown in Fig. \ref{Fig8}).

\vspace{-1em}
\section{Experiments}\label{Experiments}
We first describe the benchmark datasets and evaluation metrics (Sec. \ref{Benchmark Datasets and Evaluation Metrics}) and the implementation details (Sec. \ref{Implementation Details}). We train the proposed Diff9D using only the synthetic dataset and test it on two challenging real-world datasets to demonstrate its domain generalization ability (Sec. \ref{Evaluation on Real-World REAL275 and Wild6D Datasets}). Next, we further test Diff9D in real-world robotic grasping scenarios and deploy it on a robot to perform grasping task (Sec. \ref{Evaluation on Real-World Robotic Grasping}). Finally, we conduct some ablation studies for the condition extraction module, transformer-based denoiser, and the number of reverse diffusion time steps to explore their impact on the performance of Diff9D (Sec. \ref{Ablation Studies and Discussion}).

\begin{figure*}[htbp]
\centering
\includegraphics[width=\textwidth]{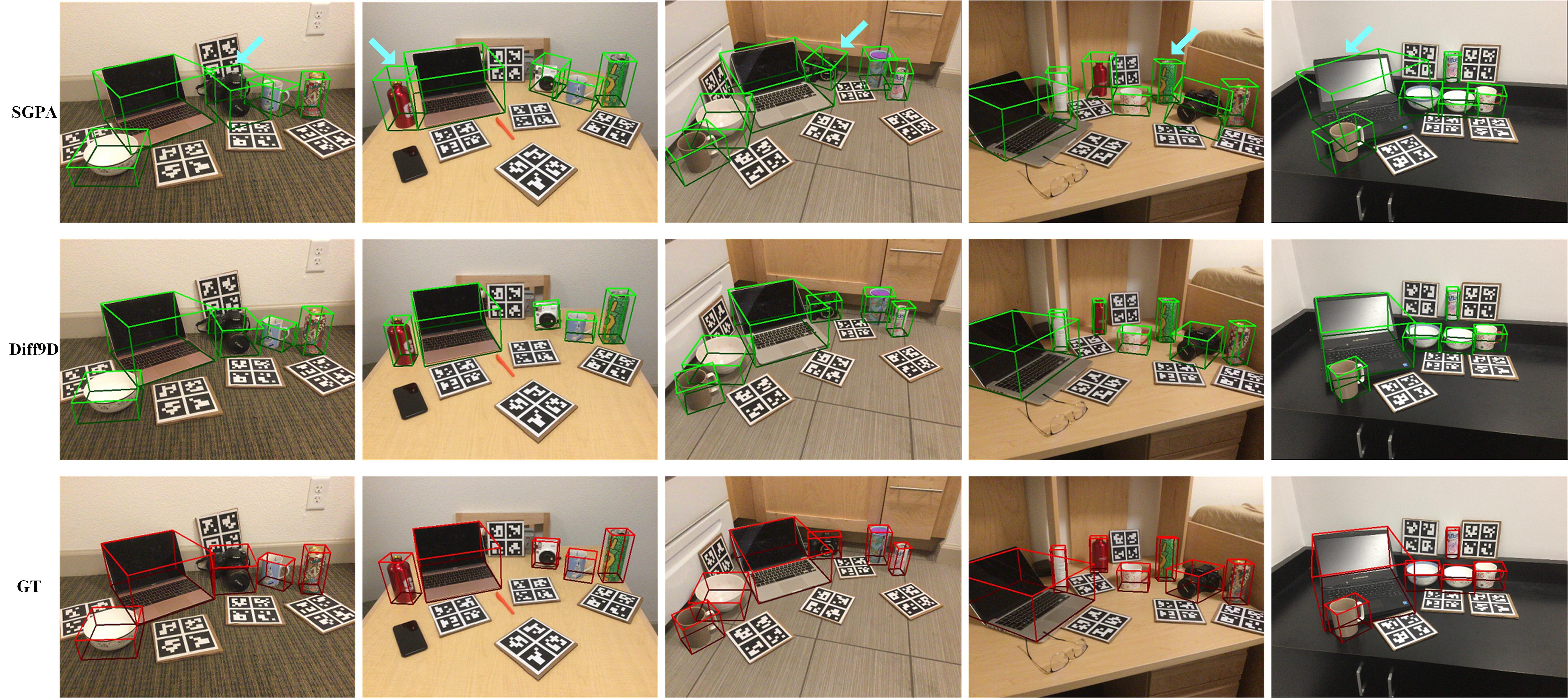}
\vspace{-2em}
\caption{Qualitative comparison results on the real-world REAL275 dataset. Both SGPA\cite{21} and Diff9D are trained using only synthetic data. For a fair comparison, all results are based on the same segmentation, i.e., by Mask R-CNN\cite{57}. The top, middle, and bottom rows denote the results of SGPA\cite{21}, Diff9D, and ground truth, respectively. Arrows point to areas of focus. We can see that Diff9D performs better than SGPA\cite{21}.}
\label{Fig9}
\vspace{-1em}
\end{figure*}

\vspace{-1em}
\subsection{Benchmark Datasets and Evaluation Metrics}\label{Benchmark Datasets and Evaluation Metrics}
\subsubsection{Benchmark Datasets}
For synthetic datasets, we choose the large CAMERA25 dataset \cite{18}, which is currently the most widely used synthetic dataset for category-level object pose estimation. For real-world datasets, the challenging and widely used REAL275 \cite{18} and Wild6D \cite{38} datasets are chosen for testing.

\noindent\emph{CAMERA25 Dataset \cite{18}} contains 275K synthetic RGB-D images for training, which includes 1085 instances from 6 categories of objects: bowl, bottle, can, camera, mug, and laptop. Note that all the 3D object models in the CAMERA25 are selected from the synthetic ShapeNet \cite{54} dataset. All RGB-D images contain multiple instances and have segmentation masks and 9-DoF pose labels.

\noindent\emph{REAL275 Dataset \cite{18}} is currently the most widely used real-world dataset for category-level object pose estimation. It contains 8K real-world RGB-D images from 18 videos. We exclusively utilize the test set of this dataset, consisting of 2754 images from 6 videos, to evaluate the performance of our proposed method. The test set includes 18 instances from 6 categories of objects, and the object categories are the same as in CAMERA25.

\noindent\emph{Wild6D Dataset \cite{38}} is a large dataset collected in the real world for evaluating self-supervised category-level object pose estimation methods. It provides annotations for only 486 test videos with different backgrounds, containing 162 objects from five categories (i.e., except ``can'' in CAMERA25 and REAL275). This paper only uses the test videos of Wild6D for experiments to enrich the real-world evaluation.

\vspace{-1em}
\subsubsection{Evaluation Metrics}
For a fair comparison with previous methods, we select the widely used 3D Intersection-over-Union ($Io{U_{3D}}$) and $n^\circ m{\rm{cm}}$ metrics for evaluation. $Io{U_{3D}}$ denotes the percentage of intersection and union of the ground-truth and the predicted 3D bounding box, which can be expressed as:
\begin{equation} \label{equation32}
Io{U_{3D}} = \frac{{{P_B} \cap {G_B}}}{{{P_B} \cup {G_B}}},
\end{equation}
where ${{G_B}}$ and ${{P_B}}$ denote the ground-truth and the predicted 3D bounding boxes, respectively. $\cap$ and $\cup$ denote the intersection and union, respectively. The predicted object pose is considered correct when the value of $Io{U_{3D}}$ is greater than a predefined threshold.
\par $n^\circ m{\rm{cm}}$ directly represents the predicted rotation and translation errors. The predicted object pose is considered correct when the rotation and translation errors are less than both $n^\circ$ and $m{\rm{cm}}$, respectively. We follow previous methods\cite{19,21,23,37,39,40} to choose 50$\%$ and 75$\%$ as the thresholds of $Io{U_{3D}}$ (termed as ${\rm{3}}{{\rm{D}}_{50}}$ and ${\rm{3}}{{\rm{D}}_{75}}$ \cite{21}) and select $5^\circ 2{\rm{cm}}$, $5^\circ 5{\rm{cm}}$, $10^\circ 2{\rm{cm}}$, and $10^\circ 5{\rm{cm}}$ for evaluation.

\vspace{-1em}
\subsection{Implementation Details}\label{Implementation Details}
Following \cite{19,20,21,23}, we set the number of points $n$ in Eq. (\ref{equation2}) and the dimension of the coordinates $k$ in Eq. (\ref{equation3}) to 1024 and 3, respectively. The diffusion time step $T$ in Eq. (\ref{equation5}) and Eq. (\ref{equation12}) is set to 1000. We take samples every 333 time steps (see Tab. \ref{table7} for reasoning) by utilizing the DDIM\cite{50} scheduler in the reverse diffusion process. The number of heads $m$ in Eq. (\ref{equation22}) is set to 16 experimentally, so the feature dimension of each head $d$ in Eq. (\ref{equation23}) is set to 112 (calculated by ${{1792} \mathord{\left/ {\vphantom {{1792} 16}} \right. \kern-\nulldelimiterspace} 16}$) \cite{55,56}. The initial RGB-D image resolution is $640 \times 480$. We first utilize Mask R-CNN \cite{57} to perform instance segmentation for the initial image, then the image is scaled to $192 \times 192$ to reduce further computation. The model weights are initialized via the default initialization method of PyTorch. The learning rate is dynamically adjusted between $1 \times 10^{-4}$ and $1 \times 10^{-6}$ through the CyclicLR function\cite{58,59}, and the step size of a cycle is set to 20K. The batch size of each step is set to 48. To avoid damaging the gripper, we rotate the grasping angle up $30^\circ $ and move the object center point up 2 centimeters as the grasping position. Experiments are conducted using an Intel Xeon Gold 6138 CPU and an NVIDIA RTX 3090 GPU.

\begin{table*}[htbp]
\renewcommand\arraystretch{1.4}
\newcommand{\tabincell}[2]{\begin{tabular}{@{}#1@{}}#2\end{tabular}}
  \centering
  \caption{Comparison on the REAL275 dataset in metrics of $Io{U_{3D}}$ ($\%$) and $n^\circ m{\rm{cm}}$ ($\%$). ``\checkmark" and ``-" indicate with and without. ``Syn", ``Real w Label", and ``Real w/o Label" indicate synthetic dataset, real-world dataset with label, and real-world dataset without label, respectively. Note that we use the $Io{U_{3D}}$ metrics of SSC6D \cite{35} and CATRE \cite{26}, which correct the small error of NOCS\cite{18} for size evaluation. The best and second-best results are bolded and underlined, respectively.}
    \vspace{-1em}
    \begin{tabular}{l|cccc|cccccc}
    \toprule[1pt]
    \multirow{2}[2]{*}{Method} & \multirow{2}[2]{*}{Syn} & \multirow{2}[2]{*}{Real w Label} & \multirow{2}[2]{*}{Real w/o Label} & \multirow{2}[2]{*}{Shape Prior} & \multicolumn{6}{c}{mean Average Precision (mAP)} \\
\cmidrule{6-11}          &       &       &    &    & ${\rm{3}}{{\rm{D}}_{50}}$ & ${\rm{3}}{{\rm{D}}_{75}}$ & \tabincell{c}{$5^\circ$$2{\rm{cm}}$} & \tabincell{c}{$5^\circ$$5{\rm{cm}}$} & \tabincell{c}{$10^\circ$$2{\rm{cm}}$} & \tabincell{c}{$10^\circ$$5{\rm{cm}}$} \\
    \midrule
    \multicolumn{11}{c}{Segmented by Mask R-CNN\cite{57} Pretrained on ImageNet Dataset} \\
    \midrule
    SPD \cite{19}  & \checkmark      & -     & -   & \checkmark  & 55.1  & 17.1  & 11.4  & 12.0  & 33.5  & 37.8  \\
    SGPA \cite{21}  & \checkmark      & -     & -  & \checkmark   & 53.1  & 17.8  & 19.8  & 27.7  & 36.5  & 62.6  \\
    STG6D \cite{23} & \checkmark      & -     & -  & \checkmark   & 48.4  & 14.4   & 20.1 & 27.9   & 39.0     & 63.7  \\
    SAR-Net \cite{34} & \checkmark      & -     & -  & \checkmark   & -  & -   & 31.6 & \underline{42.3}   & 50.3     & 68.3  \\
    SPD \cite{19} & \checkmark      & \checkmark      & -  & \checkmark   & 68.5  & 27.5  & 19.3  & 21.4  & 43.2  & 54.1  \\
    CR-Net \cite{20} & \checkmark      & \checkmark      & -  & \checkmark   & -  & 33.2  & 27.8  & 34.3  & 47.2  & 60.8  \\
    SGPA \cite{21} & \checkmark      & \checkmark      & -  & \checkmark   & 68.8  & \underline{36.6}  & \textbf{35.9}  & 39.6  & \textbf{61.3}  & \textbf{70.7}  \\
    RePoNet \cite{38} & \checkmark      & -  & \checkmark  & \checkmark  & \underline{76.0}  & -  & 29.1  & 31.3  & 48.5  & 56.8  \\
    SSC6D+ICP \cite{35} & \checkmark      & -     & \checkmark   & -   & 72.7 & - & 28.6  & 33.4  & 51.8  & 62.9  \\
    Diff9D (ours) & \checkmark      & -     & -   & -  & \textbf{76.5}  & \textbf{41.7} & \underline{35.3} & \textbf{43.9} & \underline{54.8} & \underline{70.0} \\
    \midrule
    \multicolumn{11}{c}{Segmented by Mask R-CNN\cite{57} Trained on CAMERA25 Dataset} \\
    \midrule
    DPDN \cite{39} & \checkmark      & -     & -  & \checkmark   & 67.2  & \underline{39.8}  & 29.7  & \underline{37.3}  & 53.7  & 67.0  \\
    UDA-COPE \cite{37} & \checkmark      & -     & \checkmark   & -   & \textbf{75.5}  & 34.4  & \underline{30.5}  & 34.9  & 57.0  & 66.1  \\
    TTA-COPE \cite{40} & \checkmark      & -     & -   & -   & 69.1  & 39.7  & 30.2  & 35.9  & \textbf{61.7}  & \textbf{73.2}  \\
    VI-Net \cite{32} & \checkmark      &-     & -   & -  & 33.3  & 11.3 & 11.7 & 19.9 & 14.4 & 29.1 \\
    SecondPose \cite{secondpose} & \checkmark   & -     & -   & -  & 34.9  & 14.7 & 12.7 & 21.2 & 17.6 & 34.2 \\
    Diff9D (ours) & \checkmark      & -     & -   & -  & \underline{69.2}  & \textbf{44.1} & \textbf{36.5} & \textbf{45.2} & \underline{57.7} & \underline{72.2} \\
    \midrule
    VI-Net \cite{32} & \checkmark      & \checkmark     & -   & -  & -  & 48.3 & 50.0 & \textbf{57.6} & 70.8 & \textbf{82.1} \\
    Diff9D (ours) & \checkmark      & \checkmark     & -   & -  & \textbf{79.8}  & \textbf{55.8} & \textbf{50.5} & 57.1 & \textbf{72.1} & 81.5 \\
    \bottomrule[1pt]
    \end{tabular}
  \label{table1}
  \vspace{-1em}
\end{table*}

\begin{table}[t!]
\renewcommand{\arraystretch}{1.4}
    \centering
    \caption{Comparison with diffusion model-based DiffusionNOCS \cite{DiffusionNOCS} on the REAL275 dataset. Both methods are trained solely on the synthetic CAMERA25 dataset and use RGB-D inputs.}
    \vspace{-1em}
    \begin{tabular}{p{27mm}<{\centering} | p{12mm}<{\centering} p{12mm}<{\centering} p{12mm}<{\centering}}
    \toprule[1pt]
         \multirow{2}[2]{*}{Method} & \multicolumn{3}{c}{mean Average Precision (mAP)} \\ \cmidrule{2-4}
         & $5^\circ$$5{\rm{cm}}$ & $10^\circ$$5{\rm{cm}}$ & $15^\circ$$5{\rm{cm}}$ \\
    \hline
         DiffusionNOCS \cite{DiffusionNOCS} & 35.0  & 66.6  & 77.1 \\
         Diff9D (ours) & \textbf{43.9}  &  \textbf{70.0}  &  \textbf{77.5} \\  
    \bottomrule[1pt]
    \end{tabular}
    \label{diff}
    \vspace{-1em}
\end{table}

\vspace{-0.5em}
\subsection{Evaluation on Real-World Datasets}\label{Evaluation on Real-World REAL275 and Wild6D Datasets}
\subsubsection{REAL275 Dataset}
We only use the synthetic CAMERA25 dataset to train the proposed Diff9D and compare it with one baseline method \cite{19} and nine state-of-the-art (SOTA) methods \cite{20,21,23,34,35,37,38,39,40} on the test set of the real-world REAL275 dataset. Quantitative comparison results are shown in Tab. \ref{table1}. When using Mask R-CNN pretrained on the ImageNet dataset for segmentation, Diff9D achieves 43.9$\%$ and 54.8$\%$ mean average precision (mAP) on $5^\circ$$5{\rm{cm}}$ and $10^\circ$$5{\rm{cm}}$, outperforming the baseline method SPD\cite{19} by 31.9$\%$ and 21.3$\%$, the SOTA methods SGPA \cite{21} by 16.2$\%$ and 18.3$\%$, STG6D \cite{23} by 16.0$\%$ and 15.8$\%$, respectively. In addition, Diff9D achieves 35.3$\%$ and 70.0$\%$ mAP on $5^\circ$$2{\rm{cm}}$ and $10^\circ$$2{\rm{cm}}$, outperforming the SOTA method SAR-Net \cite{34} by 3.7$\%$ and 1.7$\%$, respectively. Note that these four comparison methods all rely on shape priors, while Diff9D outperforms them without using shape priors. Moreover, SPD\cite{19}, CR-Net \cite{20}, and SGPA \cite{21} also use the real-world REAL275 dataset for training. Diff9D is 14.2$\%$, 8.5$\%$, and 5.1$\%$ better than SPD\cite{19}, CR-Net\cite{20}, and SGPA\cite{21} respectively on the 75$\%$ $Io{U_{3D}}$ metric using only synthetic dataset for training. Furthermore, we also compare with some SOTA self-supervised methods that use labeled synthetic data and unlabeled real-world data for training. Diff9D achieves 76.5$\%$ and 35.3$\%$ mAP on 50$\%$ $Io{U_{3D}}$ and $5^\circ$$2{\rm{cm}}$, outperforming RePoNet\cite{38} by 0.5$\%$ and 6.2$\%$, SSC6D+ICP \cite{35} by 3.8$\%$ and 6.7$\%$, respectively. Some qualitative results are shown in Fig. \ref{Fig9}. Additionally, we compare Diff9D with DiffusionNOCS \cite{DiffusionNOCS}, another domain-generalized category-level pose estimation method based on diffusion model. Both methods use Mask R-CNN pre-trained on the ImageNet dataset for segmentation for a fair comparison. We evaluate using the same metrics as those in DiffusionNOCS \cite{DiffusionNOCS}. The quantitative results in Tab.~\ref{diff} demonstrate that Diff9D exhibits stronger domain generalization ability compared to DiffusionNOCS \cite{DiffusionNOCS}.

\vspace{-0.5em}
\par To ensure that no real-world data is involved in any stage of the training, we use the synthetic CAMERA25 to retrain Mask R-CNN for segmentation, as shown in Tab.~\ref{table1}. When using the retrained Mask R-CNN, Diff9D achieves 69.2$\%$ and 45.2$\%$ mAP on 50$\%$ $Io{U_{3D}}$ and $5^\circ$$5{\rm{cm}}$, outperforming the SOTA self-supervised methods DPDN\cite{39} by 2.0$\%$ and 7.9$\%$, respectively. In addition, Diff9D achieves 57.7$\%$ and 72.2$\%$ mAP on $10^\circ$$2{\rm{cm}}$ and $10^\circ$$5{\rm{cm}}$, outperforming the SOTA domain adaptation method UDA-COPE \cite{37} by 0.7$\%$ and 6.1$\%$, respectively. Note that DPDN\cite{39} also uses shape priors for learning, and UDA-COPE\cite{37} requires real-world mask labels for learning, yet Diff9D outperforms these methods without using shape priors and any real-world data. Finally, Diff9D achieves 44.1$\%$ and 36.5$\%$ mAP on the most stringent 75$\%$ $Io{U_{3D}}$ and $5^\circ$$2{\rm{cm}}$ metrics, outperforming the SOTA test-time adaptation method TTA-COPE \cite{40} by 4.4$\%$ and 6.3$\%$, respectively. Moreover, we also retrain VI-Net \cite{32} and SecondPose \cite{secondpose} using only the synthetic CAMERA25 dataset, observing a significant accuracy drop in all metrics. We analyze that this is because Diff9D samples a substantial amount of object pose data along the Markov chain, leading to a more uniform pose data distribution, thus effectively reducing the domain gap between synthetic and real-world scenes. Furthermore, to assess the upper-bound performance of Diff9D, we train it using a 3:1 mix of CAMERA25 and REAL275 datasets, following the setup in VI-Net \cite{32}. The experimental results demonstrate that Diff9D achieves competitive accuracy with VI-Net \cite{32} under this mixed training configuration. The running speed of Diff9D is about 17.2 FPS, and the GPU memory occupied is about 9.6 GB.

\vspace{-1em}
\subsubsection{Wild6D Dataset}
\begin{figure*}[t!]
\vspace{-0.75em}
\centering
\includegraphics[width=0.92\textwidth]{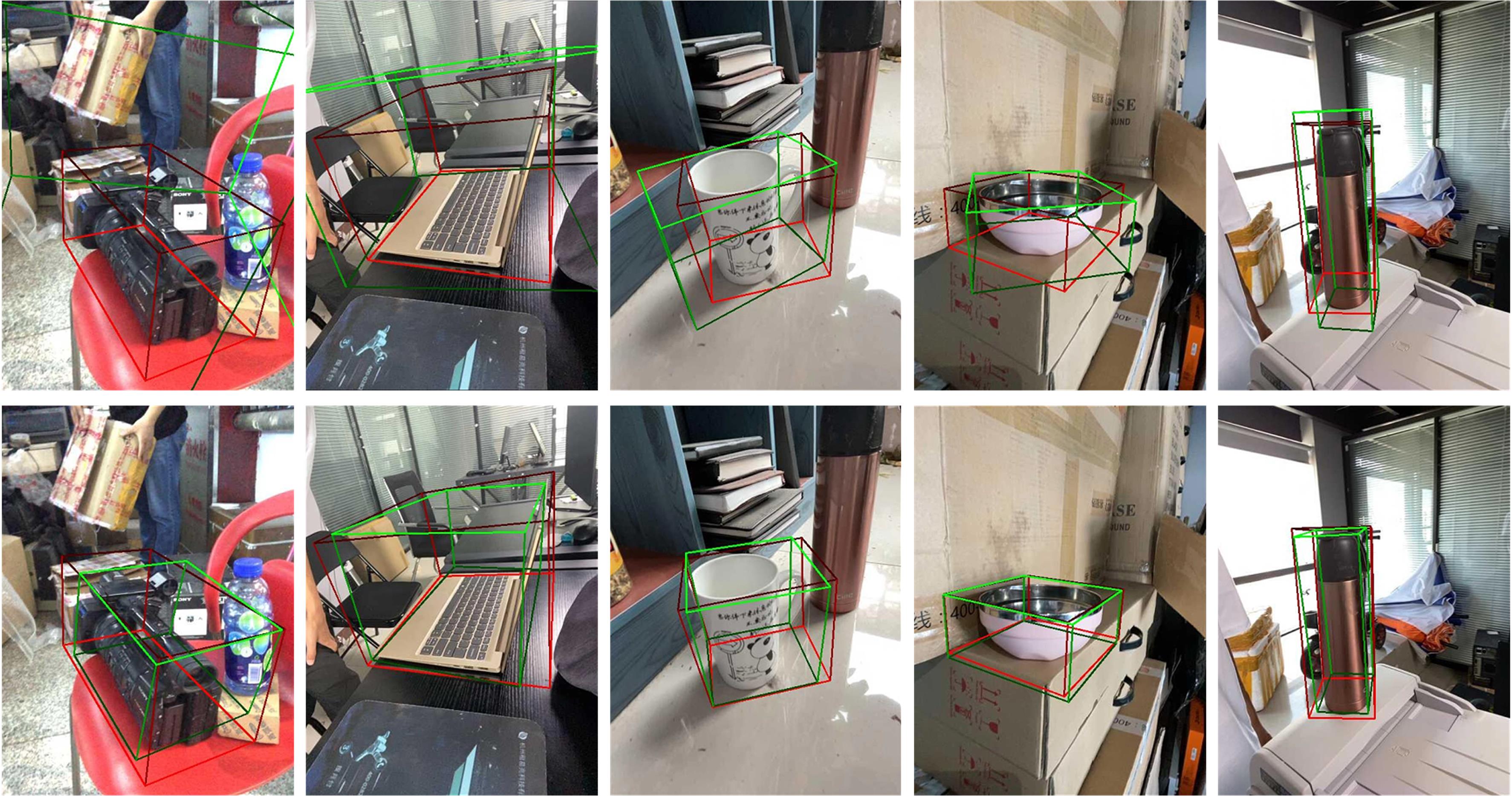}
\vspace{-0.75em}
\caption{Qualitative comparison on the real-world Wild6D dataset. SGPA\cite{21} is trained using both synthetic CAMERA25 and real-world REAL275 datasets, while Diff9D is trained using only the synthetic CAMERA25 dataset. For a fair comparison, all the results are based on the same segmentation, i.e., by Mask R-CNN\cite{57}. The top and bottom rows denote the results of SGPA\cite{21} and the proposed Diff9D, respectively. Green and Red represent the predicted and ground truth results, respectively. We can see that Diff9D performs better than SGPA\cite{21}, especially in size estimation.}
\label{Fig10}
\vspace{-1em}
\end{figure*}

To evaluate more objects and scenes in the real world beyond the REAL275 dataset, we further evaluate the proposed Diff9D and compare it with two baseline methods (NOCS \cite{18} and SPD \cite{19}) and two SOTA methods (SGPA \cite{21} and GPV-Pose \cite{29}) on the test set of the Wild6D dataset. Also note that we only use the synthetic CAMERA25 dataset to train the Diff9D, while all the comparison methods \cite{18,19,21,29} use both the synthetic CAMERA25 and real-world REAL275 datasets for training. The quantitative comparison results are shown in Tab. \ref{table2}. Specifically, Diff9D achieves 32.48$\%$ and 40.94$\%$ mAP on $10^\circ$$2{\rm{cm}}$ and $10^\circ$$5{\rm{cm}}$, outperforming the baseline methods NOCS \cite{18} by 32.44$\%$ and 40.90$\%$, and SPD \cite{19} by 12.38$\%$ and 13.10$\%$, respectively. In addition, Diff9D achieves 76.48$\%$ and 38.16$\%$ mAP on 50$\%$ and 75$\%$ $Io{U_{3D}}$, outperforming SGPA \cite{21} by 13.00$\%$ and 3.70$\%$. Furthermore, Diff9D also achieves 25.52$\%$ and 30.46$\%$ mAP on $5^\circ$$2{\rm{cm}}$ and $5^\circ$$5{\rm{cm}}$, outperforming GPV-Pose \cite{29} by 11.42$\%$ and 8.96$\%$, respectively. It is also worth noting that SPD \cite{19} and SGPA \cite{21} need to use shape priors, and GPV-Pose \cite{29} needs to use the mean size of each category as the prior, whereas Diff9D is prior-free. These results further demonstrate that Diff9D can effectively achieve the synthetic-to-real domain generalization. Some qualitative comparison results are shown in Fig. \ref{Fig10}. We find that Diff9D is significantly better than SGPA\cite{21} in size estimation.

\begin{table*}[t!]
\renewcommand\arraystretch{1.4}
\newcommand{\tabincell}[2]{\begin{tabular}{@{}#1@{}}#2\end{tabular}}
  \centering
  \caption{Comparison on the Wild6D dataset in metrics of $Io{U_{3D}}$ ($\%$) and $n^\circ m{\rm{cm}}$ ($\%$). * indicates that training with the CAMERA25 and REAL275 datasets. ``\checkmark", ``-", ``Syn", ``Real w Label", and ``Real w/o Label" represent the same meanings as explained in the caption of Tab. \ref{table1}. Note that we use the $Io{U_{3D}}$ metrics of SSC6D \cite{35} and CATRE \cite{26}, which correct the small error of NOCS\cite{18} for size evaluation.}
    \vspace{-1em}
    \begin{tabular}{l|cccc|cccccc}
    \toprule[1pt]
    \multirow{2}[2]{*}{Method} & \multirow{2}[2]{*}{Syn} & \multirow{2}[2]{*}{Real w Label} & \multirow{2}[2]{*}{Real w/o Label} & \multirow{2}[2]{*}{Shape Prior} & \multicolumn{6}{c}{mean Average Precision (mAP)} \\
\cmidrule{6-11}          &       &       &    &    & ${\rm{3}}{{\rm{D}}_{50}}$ & ${\rm{3}}{{\rm{D}}_{75}}$ & \tabincell{c}{$5^\circ$$2{\rm{cm}}$} & \tabincell{c}{$5^\circ$$5{\rm{cm}}$} & \tabincell{c}{$10^\circ$$2{\rm{cm}}$} & \tabincell{c}{$10^\circ$$5{\rm{cm}}$} \\
    \midrule
    NOCS* \cite{18} & \checkmark      & \checkmark      & -  & -   &  0.00     &  0.00     &   0.04    &   0.04    &   0.04    & 0.04  \\
    SPD* \cite{19} & \checkmark      & \checkmark      & -  & \checkmark   &  52.64  &  20.26     & 6.90   & 9.26   & 20.10   & 27.84  \\
    SGPA* \cite{21} & \checkmark      & \checkmark      & -  & \checkmark   &  \underline{63.48}   &  \underline{34.46}   &  \textbf{26.28}   &  \underline{29.18}   &  \textbf{33.80}   & 39.52  \\
    GPV-Pose* \cite{29} & \checkmark      & \checkmark      & -  & -   & -     & -     & 14.10  & 21.50  & 23.80  & \textbf{41.10}  \\
    Diff9D (ours) & \checkmark      & -     & -   & -  &  \textbf{76.48}  &  \textbf{38.16} & \underline{25.52}  & \textbf{30.46}  & \underline{32.48}  & \underline{40.94}   \\
    \bottomrule[1pt]
    \end{tabular}
  \label{table2}
\end{table*}

\vspace{-1em}
\subsection{Evaluation on Real-World Robotic Grasping}\label{Evaluation on Real-World Robotic Grasping}
\subsubsection{Real-World Robotic Grasping Scenes}
\par
\begin{figure*}[t!]
\centering
\includegraphics[width=\textwidth]{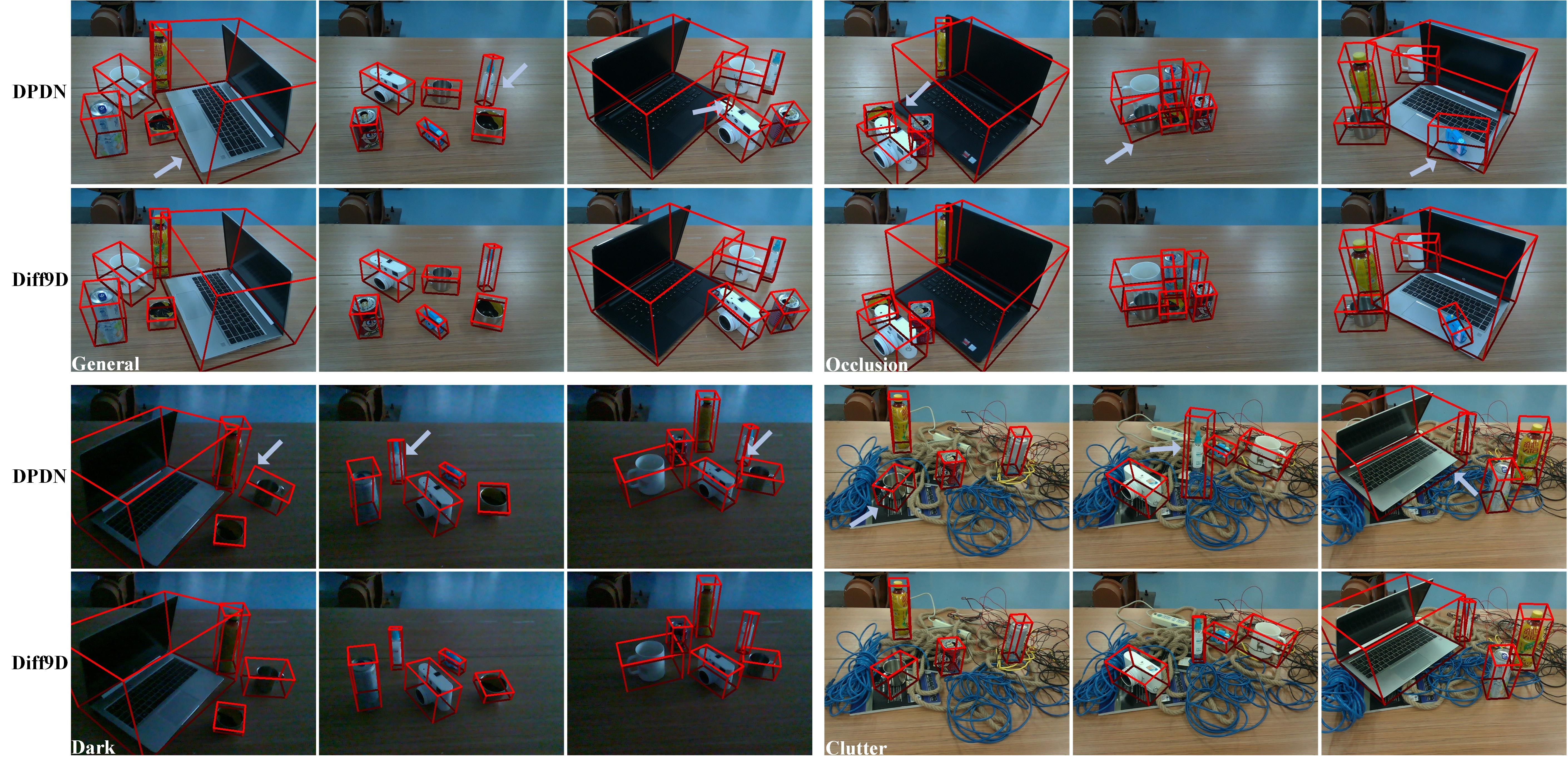}
\vspace{-2.5em}
\caption{Qualitative comparison results in four real-world robotic grasping scenes: general, occlusion, dark, and clutter. Each test scene contains four to six instances. Both DPDN\cite{39} and Diff9D are trained using only the synthetic CAMERA25 dataset. All qualitative results are based on the same segmentation results for a fair comparison, i.e., segmented by Mask R-CNN\cite{57}. Arrows point to areas of focus. We can see that Diff9D performs better than DPDN\cite{39}.}
\label{Fig11}
\vspace{-1em}
\end{figure*}

 To demonstrate the effectiveness of Diff9D for estimating the 9-DoF pose of intra-class unknown objects in real-world robotic grasping scenarios, we install an RGB-D camera on a robot and conduct tests on four scenes (general, occlusion, dark, and clutter). Specifically, we select twelve instances of six categories of objects for testing (i.e., \emph{bowl}, \emph{bottle}, \emph{can}, \emph{camera}, \emph{laptop}, and \emph{mug}), with each object category including two instances. Each test scene contained four to six instances. Some qualitative comparison results between Diff9D and DPDN\cite{39} are shown in Fig. \ref{Fig11}.

\par Note that we only use the synthetic dataset for training, so all the test instances and the robotic grasping scenes are not present in the training set. These qualitative comparison results show that Diff9D also performs better than DPDN \cite{39} in these real-world robotic grasping scenes. Specifically, DPDN has lower size and rotation estimation performance than Diff9D in some challenging scenes (i.e., occlusion and clutter). In general, Diff9D has good domain generalization ability for intra-class unknown objects and novel scenes. Also, the robustness of Diff9D is also strong in some common challenging scenes.

\vspace{-1em}
\subsubsection{Robotic Grasping}
\begin{figure}[t!]
\vspace{-1em}
\centerline{\includegraphics[width=0.97\columnwidth]{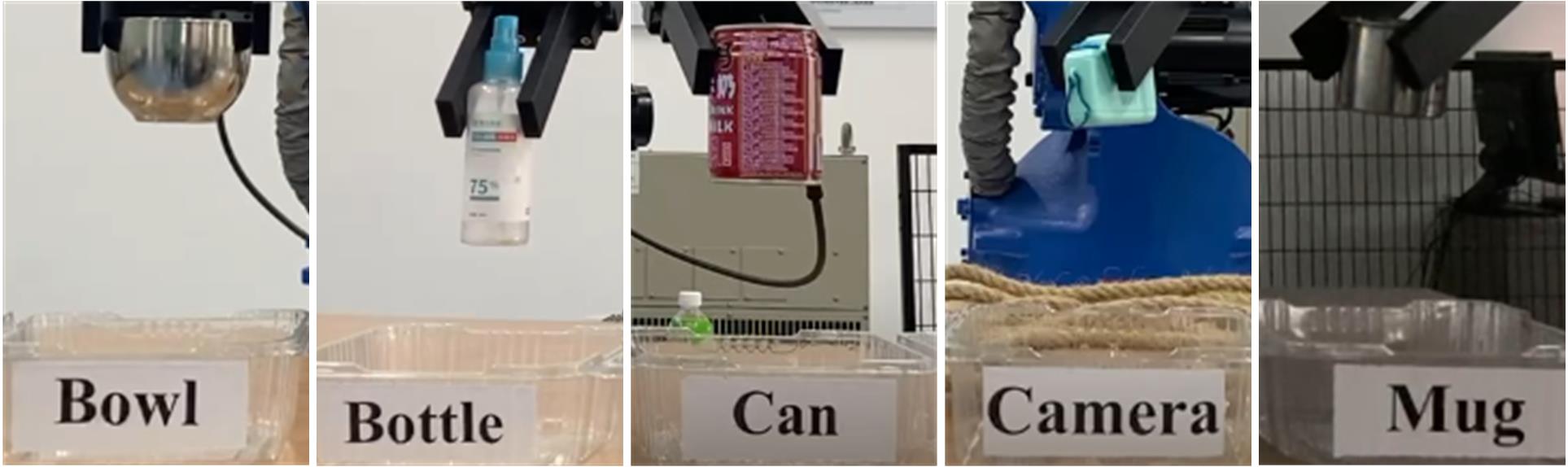}}
\vspace{-1em}
\caption{Some visualizations of the real-world robotic grasping. The grasping demo can be seen at \href{https://youtu.be/D4bV8eIUvWk}{https://youtu.be/D4bV8eIUvWk}.}
\label{Fig12}
\vspace{-1em}
\end{figure}

To demonstrate whether the proposed method can be utilized for domain-generalized robotic grasping, we develop a robotic grasping platform (as shown in Sec. \ref{Designed Robotic Grasping System}) and deploy Diff9D on it to conduct robotic grasping tests. Specifically, we conduct grasping experiments for each object category fifty times. The quantitative comparison results are shown in Tab. \ref{table3}. Note that all the results are based on the same segmentation results for a fair comparison, i.e., segmented by Mask R-CNN\cite{57}. The real-world robotic grasping experiments demonstrate the feasible generalization ability of the Diff9D, i.e., training with only rendered synthetic images can generalize to real-world robotic grasping task. Specifically, an average grasping accuracy of 80.8$\%$ can be achieved in the test of five categories of objects, outperforming STG6D \cite{23} and DPDN \cite{39} by 13.2$\%$ and 5.6$\%$, respectively. Some grasping visualizations are shown in Fig. \ref{Fig12}. The robotic grasping demo can be seen at the footnote link\footnote{\href{https://youtu.be/D4bV8eIUvWk}{https://youtu.be/D4bV8eIUvWk}}. These experimental results prove that Diff9D can be effectively utilized for robotic grasping in the real world.

\vspace{-0.5em}
\subsubsection{Limitations and Failure Cases}
\begin{figure}[t!]
\centerline{\includegraphics[width=0.85\columnwidth]{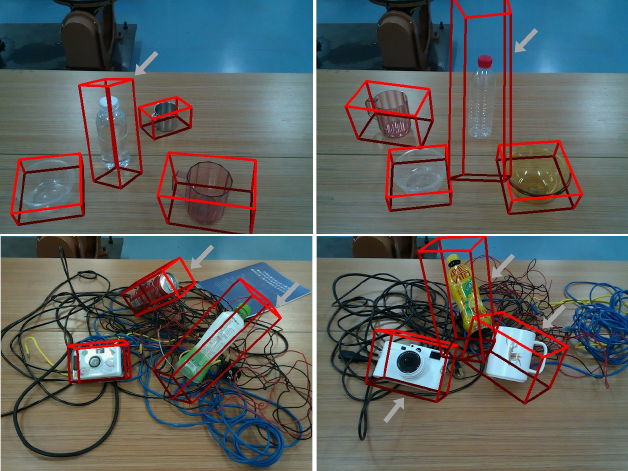}}
\vspace{-1em}
\caption{Failure cases visualization. Arrows point to focused areas.}
\label{Fig13}
\vspace{-1em}
\end{figure}

In practical experiments, we observe that pose estimation may fail when accurate geometric information for an object is difficult to obtain, such as with transparent objects (where depth imaging is challenging and texture is minimal). Additionally, pose estimation accuracy tends to decrease when objects are tilted. We attribute this to an imbalance in the training data, where most objects are upright. Figure~\ref{Fig13} presents qualitative results illustrating these observations. Based on these findings, future work will focus on enhancing the robustness of Diff9D in handling these challenging objects and scenes.

\begin{table}[t!]
\renewcommand\arraystretch{1.4}
  \centering
  \caption{The average success rate ($\%$) of robotic grasping for five categories of objects. For a fair comparison, all comparison methods are only trained on the synthetic CAMERA25 dataset. The best results are bolded.}
  \vspace{-1em}
    \begin{tabular}{p{15mm}<{\centering}|p{6mm}<{\centering}p{6mm}<{\centering}p{6mm}<{\centering}p{6mm}<{\centering}p{9mm}<{\centering}|p{7mm}<{\centering}}
    \toprule[1pt]
    Method & Bottle & Can & Bowl & Mug & Camera & Avg \\
    \midrule
    STG6D \cite{23} &  74.0  &  76.0  &  68.0  & 62.0  &  58.0  & 67.6  \\
    DPDN \cite{39} &  82.0  &  84.0  &  \textbf{80.0}  & 66.0  &  64.0  & 75.2  \\
    Diff9D &  \textbf{86.0}  &  \textbf{86.0}  &  \textbf{80.0}  & \textbf{74.0}  &  \textbf{78.0}  & \textbf{80.8}  \\
    \bottomrule[1pt]
    \end{tabular}
  \label{table3}
\end{table}

\vspace{-1em}
\subsection{Ablation Studies and Discussion}\label{Ablation Studies and Discussion}
We conduct ablation studies for the condition extraction, transformer-based denoiser, and the number of reverse diffusion time steps to explore their impact on the performance of Diff9D. We first conduct ablation studies for the input of condition extraction, which includes RGB image, point cloud, time step, and shape prior. In addition, we also perform ablation studies for the shape estimator and encoder in the condition extraction module. Then, we conduct ablation studies for the self-attention mechanism, skip connection, and the number of Transformer blocks in the transformer-based denoiser. Next, we perform ablation studies for the number of reverse diffusion time steps. Finally, we conduct ablation studies to evaluate the effectiveness of the diffusion model in reducing the domain gap between synthetic and real-world data.

\begin{table*}[t!]
\renewcommand\arraystretch{1.4}
\newcommand{\tabincell}[2]{\begin{tabular}{@{}#1@{}}#2\end{tabular}}
  \centering
  \caption{Ablation studies for the condition extraction module on the REAL275 dataset in terms of $Io{U_{3D}}$ ($\%$) and $n^\circ m{\rm{cm}}$ ($\%$). $c_{prior}$ represents using the same PointNet \cite{53} to extract global features ($1 \times 512$) of the shape prior \cite{19} and concatenate them into $c$ as a condition. ${R_s}$ and ${N_s}$ denote the shape and NOCS shape regression and encoding processes, respectively. Method 9 is the baseline method we actually use.}
  \vspace{-1em}
      \begin{tabular}{c|cccc|cc|cccccc}
    \toprule[1pt]
    \multirow{2}[2]{*}{Method} & \multirow{2}[2]{*}{$c_{rgb}$} & \multirow{2}[2]{*}{$c_{point}$} & \multirow{2}[2]{*}{$c_{timestep}$} & \multirow{2}[2]{*}{$c_{prior}$} & \multirow{2}[2]{*}{${R_s}$} & \multirow{2}[2]{*}{${N_s}$} & \multicolumn{6}{c}{mean Average Precision (mAP)} \\
\cmidrule{8-13}   &     &     &     &     &     &     & ${\rm{3}}{{\rm{D}}_{50}}$ & ${\rm{3}}{{\rm{D}}_{75}}$ & \tabincell{c}{$5^\circ$$ 2{\rm{cm}}$} & \tabincell{c}{$5^\circ$$ 5{\rm{cm}}$} & \tabincell{c}{$10^\circ$$ 2{\rm{cm}}$} & \tabincell{c}{$10^\circ$$ 5{\rm{cm}}$} \\
	\midrule
    1 & -     & \checkmark & \checkmark & - & \checkmark & \checkmark & \multicolumn{1}{c}{65.3} & \multicolumn{1}{c}{32.9} & \multicolumn{1}{c}{32.2} & \multicolumn{1}{c}{39.9} & \multicolumn{1}{c}{46.6} & \multicolumn{1}{c}{59.5} \\
    2 & \checkmark & - & \checkmark & - & \checkmark & \checkmark & \multicolumn{1}{c}{51.9} & \multicolumn{1}{c}{2.9} & \multicolumn{1}{c}{0.5} & \multicolumn{1}{c}{1.3} & \multicolumn{1}{c}{4.3} & \multicolumn{1}{c}{12.1} \\
    3 & \checkmark & \checkmark & - & - & \checkmark & \checkmark & \multicolumn{1}{c}{0.0} & \multicolumn{1}{c}{0.0} & \multicolumn{1}{c}{0.2} & \multicolumn{1}{c}{0.9} & \multicolumn{1}{c}{0.3} & \multicolumn{1}{c}{1.3} \\
    4 & \checkmark & \checkmark & \checkmark & \checkmark & \checkmark & \checkmark & \multicolumn{1}{c}{69.1} & \multicolumn{1}{c}{43.8} & \multicolumn{1}{c}{\textbf{36.6}} & \multicolumn{1}{c}{\textbf{45.5}} & \multicolumn{1}{c}{\textbf{58.1}} & \multicolumn{1}{c}{72.1} \\
    5 & \checkmark & \checkmark & \checkmark & \checkmark & - & - & \multicolumn{1}{c}{68.7} & \multicolumn{1}{c}{43.3} & \multicolumn{1}{c}{35.8} & \multicolumn{1}{c}{44.5} & \multicolumn{1}{c}{57.2} & \multicolumn{1}{c}{71.6} \\
    6 & \checkmark & \checkmark & \checkmark & - & -     & -       & \multicolumn{1}{c}{67.6} & \multicolumn{1}{c}{39.3} & \multicolumn{1}{c}{33.2} & \multicolumn{1}{c}{41.1} & \multicolumn{1}{c}{52.5} & \multicolumn{1}{c}{67.7} \\
    7 & \checkmark & \checkmark & \checkmark & - & -    & \checkmark      & \multicolumn{1}{c}{68.2} & \multicolumn{1}{c}{41.3} & \multicolumn{1}{c}{34.3} & \multicolumn{1}{c}{42.7} & \multicolumn{1}{c}{54.9} & \multicolumn{1}{c}{70.1} \\
    8 & \checkmark & \checkmark & \checkmark & - & \checkmark     & -       & \multicolumn{1}{c}{68.6} & \multicolumn{1}{c}{42.7} & \multicolumn{1}{c}{35.4} & \multicolumn{1}{c}{43.9} & \multicolumn{1}{c}{56.3} & \multicolumn{1}{c}{71.0} \\
    9 & \checkmark & \checkmark & \checkmark & - & \checkmark & \checkmark & \multicolumn{1}{c}{\textbf{69.2}} & \multicolumn{1}{c}{\textbf{44.1}} & \multicolumn{1}{c}{36.5} & \multicolumn{1}{c}{45.2} & \multicolumn{1}{c}{57.7} & \multicolumn{1}{c}{\textbf{72.2}} \\
    \bottomrule[1pt]
    \end{tabular}
  \label{table4}
  \vspace{-1em}
\end{table*}

\vspace{-1em}
\subsubsection{Ablation Studies for Condition Extraction}
\noindent\emph{1) Input Data:} First, we conduct ablation studies on each input of the condition extraction module. When the input RGB image is removed, the experimental results, shown in the first row of Tab. \ref{table4}, indicate that the 75$\%$ $Io{U_{3D}}$ and $5^\circ$$2{\rm{cm}}$ metrics drop by 11.2 $\%$ and 4.3 $\%$, respectively. Since RGB images can provide texture features of the object, including them as a condition provides effective guidance to the object pose diffusion process. When the point cloud and time step conditions are removed, the experimental results (second and third rows of Tab. \ref{table4}) reveal significant drops in all metrics. It is well-known that point cloud is the most crucial input for object pose estimation, as it provides positional and geometric information about the object. Also, the time step is a critical condition for the diffusion model, controlling the extent of noising/denoising. Hence, the point cloud and time step conditions are both essential.
\par Moreover, considering that many comparison methods are shape prior-based \cite{19,20,21,23,38}, we also use the same PointNet \cite{53} to extract global features of the shape prior as a condition. The experimental results, shown in the fourth row of Tab. \ref{table4}, indicate that the performance with the shape prior condition is similar to that without it (the ninth row). We argue that this is because the shape estimator and encoder already introduce global geometric features. Hence, the shape prior does not provide additional performance gains. To validate this hypothesis, we retain the shape prior as a condition while removing the shape estimator and encoder. The experimental results, shown in the fifth row of Tab. \ref{table4}, demonstrate that the shape prior indeed brings performance gains to the pose diffusion process (compared to the sixth row of Tab. \ref{table4}).

\begin{table}[t!]
\vspace{-0.5em}
\renewcommand\arraystretch{1.4}
\newcommand{\tabincell}[2]{\begin{tabular}{@{}#1@{}}#2\end{tabular}}
  \centering
  \caption{Ablation studies for the self-attention mechanism and skip connection in the transformer-based denoiser on the REAL275 dataset in terms of $Io{U_{3D}}$ ($\%$) and $n^\circ m{\rm{cm}}$ ($\%$). \emph{SA} and \emph{SC} are the abbreviations of self-attention and skip connection, respectively. * represents using cross-attention mechanism instead of self-attention mechanism.}
  \vspace{-1em}
      \begin{tabular}{p{4.5mm}<{\centering} p{4.5mm}<{\centering} | p{5mm}<{\centering} p{5mm}<{\centering} p{7.5mm}<{\centering} p{7.5mm}<{\centering} p{7.5mm}<{\centering} p{8.5mm}<{\centering}}
    \toprule[1pt]
    \multirow{2}[2]{*}{${SA}$} & \multirow{2}[2]{*}{${SC}$} & \multicolumn{6}{c}{mean Average Precision (mAP)} \\
\cmidrule{3-8}          &       & ${\rm{3}}{{\rm{D}}_{50}}$ & ${\rm{3}}{{\rm{D}}_{75}}$ & \tabincell{c}{$5^\circ$$ 2{\rm{cm}}$} & \tabincell{c}{$5^\circ$$ 5{\rm{cm}}$} & \tabincell{c}{$10^\circ$$ 2{\rm{cm}}$} & \tabincell{c}{$10^\circ$$ 5{\rm{cm}}$} \\
	\midrule
    *     & \checkmark      & \multicolumn{1}{c}{52.5} & \multicolumn{1}{c}{16.7} & \multicolumn{1}{c}{20.2} & \multicolumn{1}{c}{27.1} & \multicolumn{1}{c}{37.6} & \multicolumn{1}{c}{63.1} \\
    \checkmark     & -       & \multicolumn{1}{c}{65.4} & \multicolumn{1}{c}{41.2} & \multicolumn{1}{c}{32.7} & \multicolumn{1}{c}{40.9} & \multicolumn{1}{c}{54.2} & \multicolumn{1}{c}{68.5} \\
    \checkmark     & \checkmark      & \multicolumn{1}{c}{\textbf{69.2}} & \multicolumn{1}{c}{\textbf{44.1}} & \multicolumn{1}{c}{\textbf{36.5}} & \multicolumn{1}{c}{\textbf{45.2}} & \multicolumn{1}{c}{\textbf{57.7}} & \multicolumn{1}{c}{\textbf{72.2}} \\
    \bottomrule[1pt]
    \end{tabular}
  \label{table5}
  \vspace{-1em}
\end{table}

\noindent\emph{2) Shape Estimator and Encoder:} First, we remove the shape estimator and encoder in the condition extraction module. The quantitative results are shown in the sixth row of Tab. \ref{table4}. Compared with the original Diff9D, the 50$\%$ and 75$\%$ $Io{U_{3D}}$ metrics drop by 1.6$\%$ and 4.8$\%$, respectively. Then, we just remove the shape regression and encoding processes (${R_s}$) in the shape estimator and encoder. Compared with the original Diff9D, the $5^\circ$$2{\rm{cm}}$ and $5^\circ$$5{\rm{cm}}$ metrics drop by 2.2$\%$ and 2.5$\%$, respectively. The quantitative results are shown in the seventh row of Tab. \ref{table4}. Next, we simply remove the NOCS shape regression and encoding modules (${N_s}$) in the shape estimator and encoder. Compared with the original Diff9D, the $10^\circ$$2{\rm{cm}}$ and $10^\circ$$5{\rm{cm}}$ metrics drop by 1.4$\%$ and 1.2$\%$, respectively. Results are shown in the eighth row of Tab. \ref{table4}. These results show that both the shape and NOCS shape regression and encoding processes in the condition extraction module contribute to enhancing the accuracy of Diff9D. Specifically, these two processes introduce inherent 3D geometric information, providing geometric guidance to the pose diffusion process.

\begin{table}[t!]
\vspace{-0.5em}
\renewcommand\arraystretch{1.4}
\newcommand{\tabincell}[2]{\begin{tabular}{@{}#1@{}}#2\end{tabular}}
  \centering
  \caption{Ablation studies for the number of transformer blocks (${T_b}$) in the transformer-based denoiser on the REAL275 dataset in terms of $Io{U_{3D}}$ ($\%$) and $n^\circ m{\rm{cm}}$ ($\%$). \emph{SC} is the abbreviation of skip connection. This paper finally sets ${T_b}=7$ and $SC=3$.}
  \vspace{-1em}
      \begin{tabular}{p{4.5mm}<{\centering} p{4.5mm}<{\centering} | p{5mm}<{\centering} p{5mm}<{\centering} p{7.5mm}<{\centering} p{7.5mm}<{\centering} p{7.5mm}<{\centering} p{8.5mm}<{\centering}}
    \toprule[1pt]
    \multirow{2}[2]{*}{${T_b}$} & \multirow{2}[2]{*}{$SC$} & \multicolumn{6}{c}{mean Average Precision (mAP)} \\
\cmidrule{3-8}          &       & ${\rm{3}}{{\rm{D}}_{50}}$ & ${\rm{3}}{{\rm{D}}_{75}}$ & \tabincell{c}{$5^\circ$$ 2{\rm{cm}}$} & \tabincell{c}{$5^\circ$$ 5{\rm{cm}}$} & \tabincell{c}{$10^\circ$$ 2{\rm{cm}}$} & \tabincell{c}{$10^\circ$$ 5{\rm{cm}}$} \\
	\midrule
    3     & 1       & \multicolumn{1}{c}{65.1} & \multicolumn{1}{c}{41.3} & \multicolumn{1}{c}{32.2} & \multicolumn{1}{c}{41.1} & \multicolumn{1}{c}{54.5} & \multicolumn{1}{c}{68.7} \\
    5     & 2      & \multicolumn{1}{c}{68.2} & \multicolumn{1}{c}{42.9} & \multicolumn{1}{c}{34.5} & \multicolumn{1}{c}{43.4} & \multicolumn{1}{c}{56.2} & \multicolumn{1}{c}{71.1} \\
    7     & 3      & \multicolumn{1}{c}{69.2} & \multicolumn{1}{c}{\textbf{44.1}} & \multicolumn{1}{c}{\textbf{36.5}} & \multicolumn{1}{c}{45.2} & \multicolumn{1}{c}{57.7} & \multicolumn{1}{c}{\textbf{72.2}} \\
    9     & 4       & \multicolumn{1}{c}{\textbf{69.3}} & \multicolumn{1}{c}{44.0} & \multicolumn{1}{c}{\textbf{36.5}} & \multicolumn{1}{c}{\textbf{45.3}} & \multicolumn{1}{c}{57.7} & \multicolumn{1}{c}{\textbf{72.2}} \\
    11     & 5       & \multicolumn{1}{c}{\textbf{69.3}} & \multicolumn{1}{c}{\textbf{44.1}} & \multicolumn{1}{c}{36.3} & \multicolumn{1}{c}{45.1} & \multicolumn{1}{c}{\textbf{57.9}} & \multicolumn{1}{c}{\textbf{72.2}} \\
    \bottomrule[1pt]
    \end{tabular}
  \label{table6}
  \vspace{-1em}
\end{table}

\vspace{-0.5em}
\subsubsection{Ablation Studies for the Self-Attention Mechanism and Skip Connection in the Transformer-Based Denoiser}
First, we present the results of directly using the commonly used cross-attention mechanism\cite{60} instead of our proposed self-attention mechanism in the first row of Tab. \ref{table5}. It is evident that there is a substantial decline in all evaluation metrics. Specifically, if the cross-attention mechanism is applied to the conditions and the diffusion target (object pose), it will require upscaling the object pose features to the same dimension as the conditions. However, due to the sparsity of the object pose, this is redundant. Then, we remove the skip connections in the transformer-based denoiser. The quantitative results are shown in the second row of Tab. \ref{table5}. Compared with the original Diff9D, the 50$\%$ and 75$\%$ $Io{U_{3D}}$ metrics drop by 3.8$\%$ and 2.9$\%$, respectively. Since skip connections can retain more spatial information during diffusion, they play a crucial role in the denoising process.

\vspace{-0.5em}
\subsubsection{Ablation Studies for the Number of Transformer Blocks in the Transformer-Based Denoiser}
First, we set the number of Transformer blocks (${T_b}$) and skip connections ($SC$) to 3 and 1, respectively. Compared with the original Diff9D, the $5^\circ$$2{\rm{cm}}$ metric drops from 36.5$\%$ to 32.2$\%$, and the $5^\circ$$5{\rm{cm}}$ metric drops from 45.2$\%$ to 41.1$\%$. The quantitative results are shown in the first row of Tab. \ref{table6}. Next, we set ${T_b}$ and $SC$ to 5 and 2, respectively. Compared with the original Diff9D, the $10^\circ$$2{\rm{cm}}$ metric drops from 57.7$\%$ to 56.2$\%$, and the $10^\circ$$5{\rm{cm}}$ metric drops from 72.2$\%$ to 71.1$\%$. Results are shown in the second row of Tab. \ref{table6}. Furthermore, we also increase ${T_b}$ to 9 and 11, so $SC$ corresponds to 4 and 5. Results are shown in the fourth and fifth rows of Tab. \ref{table6}. From these experimental results, we can see that when ${T_b}=7$ and $SC=3$, the transformer-based denoiser performance reaches saturation.

\begin{table}[t!]
\renewcommand\arraystretch{1.4}
\newcommand{\tabincell}[2]{\begin{tabular}{@{}#1@{}}#2\end{tabular}}
  \centering
  \caption{Ablation studies for the number of reverse diffusion time steps and the schedulers on the REAL275 dataset in terms of $Io{U_{3D}}$ ($\%$), $n^\circ m{\rm{cm}}$ ($\%$), and running speed (FPS). This paper finally sets $S=3$. * denotes using DPM-Solver++ scheduler \cite{DPM-Solver++} instead of DDIM scheduler \cite{50}.}
  \vspace{-1em}
      \begin{tabular}{p{5mm}<{\centering} | p{5mm}<{\centering} p{5mm}<{\centering} p{7.5mm}<{\centering} p{7.5mm}<{\centering} p{7.5mm}<{\centering} p{9mm}<{\centering} | p{7.5mm}<{\centering}}
    \toprule[1pt]
    \multirow{2}[2]{*}{$S$} & \multicolumn{6}{c|}{mean Average Precision (mAP)} & \multirow{2}[2]{*}{Speed} \\
\cmidrule{2-7}          & ${\rm{3}}{{\rm{D}}_{50}}$ & ${\rm{3}}{{\rm{D}}_{75}}$ & \tabincell{c}{$5^\circ$$ 2{\rm{cm}}$} & \tabincell{c}{$5^\circ$$ 5{\rm{cm}}$} & \tabincell{c}{$10^\circ$$ 2{\rm{cm}}$} & \tabincell{c}{$10^\circ$$ 5{\rm{cm}}$} &  \\
    \midrule
    1     &  0.0    &  0.0     &  0.0    &  0.0   &  0.0     &  0.0   & 32.3 \\
    2     &  68.8     &  44.0     &  36.4    & 44.7    &  57.5     &  72.0   & 22.2 \\
    3     &  69.2     &  44.1     &  \textbf{36.5}    & 45.2    &  57.7     &  72.2   & 17.2 \\
    3*     &  69.3     &  44.1     &  36.3    & \textbf{45.3}    &  57.7     &  72.1   & 17.4 \\
    4     &  69.2     &  44.2     &  36.4    & 45.1    &  57.7     &  72.3   & 14.1 \\
    5     &  69.2     &  44.3     &  \textbf{36.5}    & \textbf{45.3}    &  57.7     &  72.2   & 11.8 \\
    10     &  69.2     &  \textbf{44.5}     &  \textbf{36.5}    & 45.2    &  \textbf{57.8}     &  \textbf{72.4}   & 6.5 \\
    20     & 69.3      & \textbf{44.5}       & \textbf{36.5}    & \textbf{45.3}    & 57.6      & 72.0    & 3.5 \\
    50     &  69.3     &  44.4     &  36.4     &  44.9     &  57.6     &   71.9    &  1.1 \\
    100     &   \textbf{69.4}    &   44.3    &   36.1    &   44.5    &   57.5    &  71.7     & 0.7 \\
    200     &   \textbf{69.4}    &   44.2    &   35.9    &   44.3    &   57.3    &  71.5     & 0.4 \\
    1000     &   69.2    &   44.3    &   35.8    &   44.1    &   57.4    &  71.5     & 0.1 \\
    \bottomrule[1pt]
    \end{tabular}%
  \label{table7}%
  \vspace{-1em}
\end{table}%

\vspace{-0.5em}
\subsubsection{Ablation Studies for the Number of Reverse Diffusion Time Steps and the Schedulers}
To improve the speed of Diff9D, we utilize the DDIM scheduler \cite{50} to sample the reverse diffusion process. We show that only as few as 3 steps are sufficient to recover the 9-DoF object pose from Gaussian noise using our proposed Diff9D model. The sampling time steps affect both the speed and accuracy of pose diffusion. We conduct ten sets of ablation experiments to explore the effect of the sample time steps. Specifically, we set the sampling time steps $S$ to 1, 2, 3, 4, 5, 10, 20, 50, 100, 200, and 1000 (original DDPM \cite{49}). Results are shown in Tab. \ref{table7}. When $S=1$, all evaluation metrics are 0, which proves that Gaussian noise cannot be directly diffused to object pose through a single time step. Next, when set $S=2$, the pose estimation accuracy is greatly improved and is close to the optimal accuracy. Then gradually increasing $S$, we find that the pose estimation accuracy slowly improves and reaches an optimum value at $S=10$. Beyond 10, there is a gradual decline in pose estimation accuracy. Given the sparsity of object pose data (only 15 values), we argue that the proposed Diff9D differs from other dense diffusion tasks (e.g., image generation) and can be efficient with fewer diffusion steps. Additionally, the dense denoising of sparse object pose data may introduce extraneous noise, resulting in a slight decrease of accuracy. Considering both accuracy and speed, we set $S=3$ beyond which the improvement is minimal. Using such a small number of steps allows us to achieve near real-time robotic grasping at 17.2 FPS. Moreover, we also use the popular scheduler DPM-Solver++ \cite{DPM-Solver++} as an alternative. The experimental results are shown in the fourth row of Tab. \ref{table7}. We can see that its performance is similar to that of DDIM.

\begin{table}[t!]
\renewcommand\arraystretch{1.4}
\newcommand{\tabincell}[2]{\begin{tabular}{@{}#1@{}}#2\end{tabular}}
  \centering
  \caption{Ablation studies for the effectiveness of diffusion model on the REAL275 dataset in terms of $Io{U_{3D}}$ ($\%$) and $n^\circ m{\rm{cm}}$ ($\%$).}
  \vspace{-1em}
      \begin{tabular}{p{17mm}<{\centering} | p{5mm}<{\centering} p{5mm}<{\centering} p{7.5mm}<{\centering} p{7.5mm}<{\centering} p{7.5mm}<{\centering} p{8.5mm}<{\centering}}
    \toprule[1pt]
    \multirow{2}[2]{*}{Method} & \multicolumn{6}{c}{mean Average Precision (mAP)} \\
\cmidrule{2-7} & ${\rm{3}}{{\rm{D}}_{50}}$ & ${\rm{3}}{{\rm{D}}_{75}}$ & \tabincell{c}{$5^\circ$$ 2{\rm{cm}}$} & \tabincell{c}{$5^\circ$$ 5{\rm{cm}}$} & \tabincell{c}{$10^\circ$$ 2{\rm{cm}}$} & \tabincell{c}{$10^\circ$$ 5{\rm{cm}}$} \\
	\midrule
    w/o Diff + 1 & \multicolumn{1}{c}{62.9} & \multicolumn{1}{c}{38.4} & \multicolumn{1}{c}{29.2} & \multicolumn{1}{c}{38.7} & \multicolumn{1}{c}{46.8} & \multicolumn{1}{c}{70.0} \\
    w/o Diff + 2 & \multicolumn{1}{c}{63.7} & \multicolumn{1}{c}{39.2} & \multicolumn{1}{c}{30.1} & \multicolumn{1}{c}{39.5} & \multicolumn{1}{c}{47.2} & \multicolumn{1}{c}{71.0} \\
    w/o Diff + 3 & \multicolumn{1}{c}{65.5} & \multicolumn{1}{c}{39.4} & \multicolumn{1}{c}{30.7} & \multicolumn{1}{c}{39.8} & \multicolumn{1}{c}{47.9} & \multicolumn{1}{c}{71.4} \\
    Diff9D (+ 2)       & \multicolumn{1}{c}{68.8} & \multicolumn{1}{c}{44.0} & \multicolumn{1}{c}{36.4} & \multicolumn{1}{c}{44.7} & \multicolumn{1}{c}{57.5} & \multicolumn{1}{c}{72.0} \\
    Diff9D (+ 3)       & \multicolumn{1}{c}{\textbf{69.2}} & \multicolumn{1}{c}{\textbf{44.1}} & \multicolumn{1}{c}{\textbf{36.5}} & \multicolumn{1}{c}{\textbf{45.2}} & \multicolumn{1}{c}{\textbf{57.7}} & \multicolumn{1}{c}{\textbf{72.2}} \\
    \bottomrule[1pt]
    \end{tabular}
  \label{table8}
  \vspace{-1em}
\end{table}

\vspace{-0.5em}
\subsubsection{Ablation Studies for Effectiveness of Diffusion Model}
To demonstrate the effectiveness of the diffusion model in reducing the domain gap between synthetic and real-world data, we remove the diffusion model components (both noising and denoising) from Diff9D, transforming it into a plain regression model. Overall, to ensure a fair comparison, we adopt the condition extraction module of Diff9D for feature encoding, utilize a network with the same structure as the transformer-based denoiser of Diff9D for feature fusion, and employ an identical pose decoding approach as Diff9D. Specifically, we extract features from the RGB image and point cloud in the same manner, and similarly utilize the shape estimator and encoder to extract shape features. These features are then concatenated and fed into the same transformer-based denoiser network for feature fusion. Finally, the fused features are input into fully connected layers for decoupled pose regression (each of ${t_p}$, $R$, and $s$ is regressed by three fully connected layers). For a further fair comparison, we also iterate the transformer-based denoiser network 2 and 3 times for feature fusion. The experimental results, shown in the first three rows of Tab. \ref{table8}, clearly indicate that all metrics drop when the diffusion model is removed. We attribute this to two main reasons: \textbf{1)} Since the sampling of object pose data on the Markov chain is removed, the pose data distribution is limited in the synthetic domain, so the performance gain of domain generalization cannot be obtained. Conversely, Diff9D samples a large amount of object pose data on the Markov chain based on the denoising diffusion probabilistic model, making the data distribution more uniform \cite{Ed-sam}, thus contributing to reducing the domain gap between synthesis and the real world. \textbf{2)} We only use lightweight feature extraction backbone networks (ResNet18 \cite{52} and PointNet \cite{53}) and direct feature concatenation, and only utilize the global features of RGB and point cloud images, while the plain regression models typically rely more on sufficient feature extraction and fusion. In contrast, image features are just conditions for the diffusion model, so Diff9D relies less on fine features than the regression model. Overall, the experimental results prove that the diffusion model has a stronger domain generalization capability compared to the plain regression model.

\vspace{-1em}
\section{Conclusion}\label{Conclusion}
We presented a DDPM-based method for domain-generalized category-level 9-DoF object pose estimation. 
The proposed Diff9D design is motivated by the latent generalization ability of the diffusion model to solve the domain generalization problem in object pose estimation. We showed that reverse diffusion can be performed in as few as 3 steps to achieve state-of-the-art accuracy at near real-time performance for robotic grasping tasks.
Our model does not require real-world training data nor object shape priors, yet it generalizes well to real-world scenarios. Extensive experiments on two benchmark datasets and deployment on a robotic arm show that our model achieves the overall best accuracy compared to existing methods.

\vspace{-1em}
\ifCLASSOPTIONcompsoc
  \section*{Acknowledgments}
\else
  \section*{Acknowledgment}
\fi

This work was supported by the National Natural Science Foundation of China under Grant U22A2059 and Grant 62473141, China Scholarship Council under Grant 202306130074 and Grant 202206130048, Natural Science Foundation of Hunan Province under Grant 2024JJ5098, and by the Open Foundation of the State Key Laboratory of Advanced Design and Manufacturing for Vehicle Body and the Engineering Research Center of Multi-Mode Control Technology and Application for Intelligent System of the Ministry of Education. Ajmal Mian was supported by the Australian Research Council Future Fellowship Award funded by the Australian Government under Project FT210100268.

\ifCLASSOPTIONcaptionsoff
  \newpage
\fi

\vspace{-4em}
\begin{IEEEbiography}[{\includegraphics[width=1in,height=1.25in,clip,keepaspectratio]{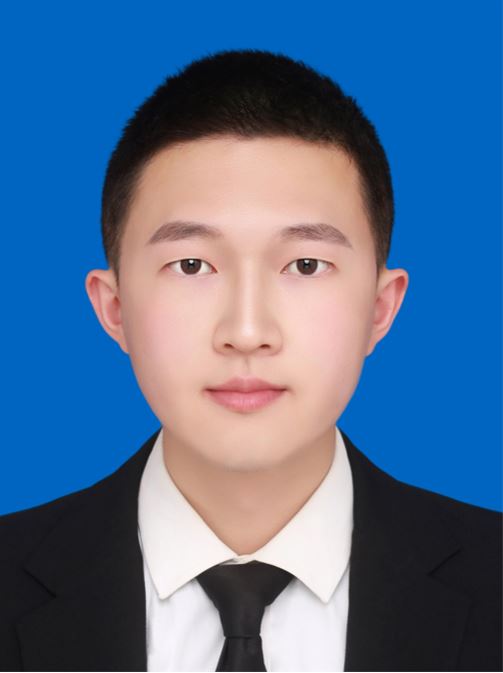}}]{Jian Liu} is currently pursuing his Ph.D. degree at the National Engineering Research Center of Robot Visual Perception and Control Technology, College of Electrical and Information Engineering, Hunan University, Changsha, China, under the supervision of Prof. Wei Sun. He is currently a Visiting Scholar at the Department of Computer Science of the University of Western Australia, Perth, WA, Australia, under the supervision of Prof. Ajmal Mian. His current research interests include 3D computer vision, object pose estimation, and robotic manipulation. He served as a reviewer for more than ten journals and conferences, including the IEEE TPAMI, IEEE TIP, IEEE TNNLS, IEEE TII, IEEE TCSVT, and IEEE ICRA, etc.
\end{IEEEbiography}

\begin{IEEEbiography}[{\includegraphics[width=1in,height=1.25in,clip,keepaspectratio]{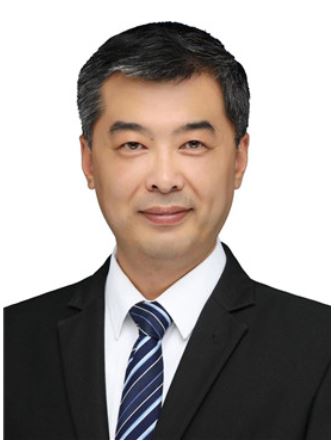}}]{Wei Sun}
received the B.E. degree in industrial automation from Hunan University, Changsha, China, in 1997, the M.E., and Ph.D. degrees in control theory and control engineering from Hunan University, Changsha, China, in 1999, and 2003, respectively. He is currently a Full Professor at Hunan University and the Chief Scientist of the National Engineering Research Center of Robot Visual Perception and Control Technology.
\par Prof. Sun received one Second-Grade National Technology Invention Award of China in 2018 and two Second-Grade National Science and Technology Progress Awards of China in 2004 and 2006. His research interests include computer vision, robotics, and artificial intelligence, with over 200 publications in these areas.
\end{IEEEbiography}

\begin{IEEEbiography}[{\includegraphics[width=1in,height=1.25in,clip,keepaspectratio]{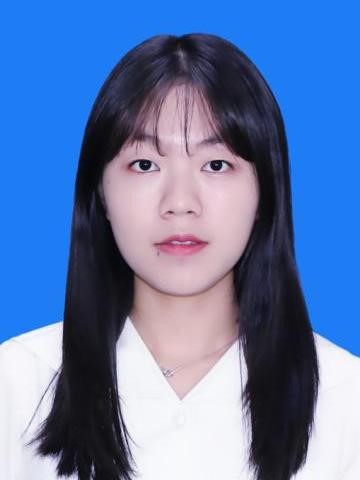}}]{Hui Yang}
received the B.S. degree from Dalian Maritime University, Dalian, China, in 2022. She is currently pursuing his Ph.D. degree at the National Engineering Research Center of Robot Visual Perception and Control Technology, College of Electrical and Information Engineering, Hunan University, Changsha, China, under the supervision of Prof. Wei Sun. Her research interests include 6D object pose estimation, point cloud analysis, and robotic manipulation.
\end{IEEEbiography}

\begin{IEEEbiography}[{\includegraphics[width=1in,height=1.25in,clip,keepaspectratio]{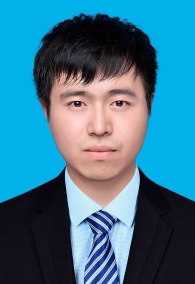}}]{Pengchao Deng}
received the M.S degree in Software Engineering from Xi'an Jiaotong University, Xi'an, China, in 2017. He is currently pursuing his Ph.D. degree at the Institute of Artificial Intelligence and Robotics, Xi'an Jiaotong University. He is currently a Visiting Scholar with the University of Western Australia, Perth, WA, Australia, under the supervision of Prof. Mohammed Bennamoun. His research interests focus on 3D ranging, domain generation, etc.
\end{IEEEbiography}

\begin{IEEEbiography}[{\includegraphics[width=1in,height=1.25in,clip,keepaspectratio]{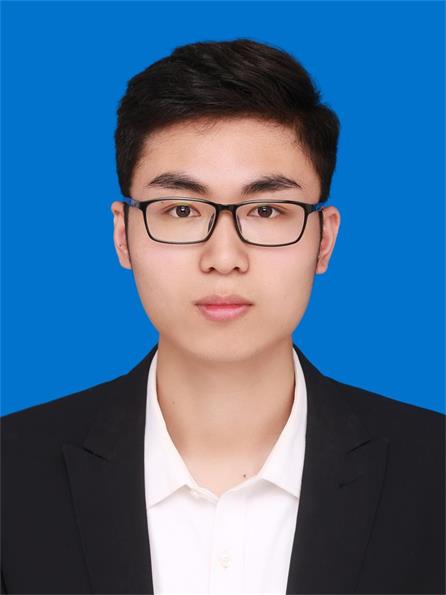}}]{Chongpei Liu}
is currently pursuing his Ph.D. degree at the National Engineering Research Center of Robot Visual Perception and Control Technology, College of Electrical and Information Engineering, Hunan University, Changsha, China. He is currently a Visiting Scholar at the University of Trento, Italy, under the supervision of Prof. Nicu Sebe. His current research interests include 3D computer vision, deep learning, and 6D object pose estimation.
\end{IEEEbiography}

\begin{IEEEbiography}[{\includegraphics[width=1in,height=1.25in,clip,keepaspectratio]{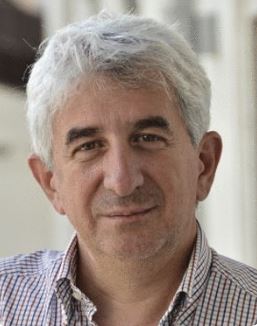}}]{Nicu Sebe (Senior Member, IEEE)}
is a Professor at the University of Trento, Italy, leading the research in the areas of multimedia information retrieval and human behavior understanding. He was the General Co-Chair of ACM Multimedia 2013 and 2022, and the Program Chair of ACM Multimedia 2007 and 2011, ECCV 2016, ICCV 2017, and ICPR 2020.
\par Prof. Sebe is a fellow of the International Association for Pattern Recognition (IAPR). He is an Associate Editor of the IEEE \textsc{Transactions on Pattern Analysis and Machine Intelligence} (TPAMI). He also served as a Guest Editor for the \textit{International Journal of Computer Vision} (IJCV), IEEE \textsc{Transactions on Multimedia} (TMM), the IEEE \textsc{Transactions on Big Data} (TBD), and the \textit{Computer Vision and Image Understanding} (CVIU) journal, etc.
\end{IEEEbiography}

\begin{IEEEbiography}[{\includegraphics[width=1in,height=1.25in,clip,keepaspectratio]{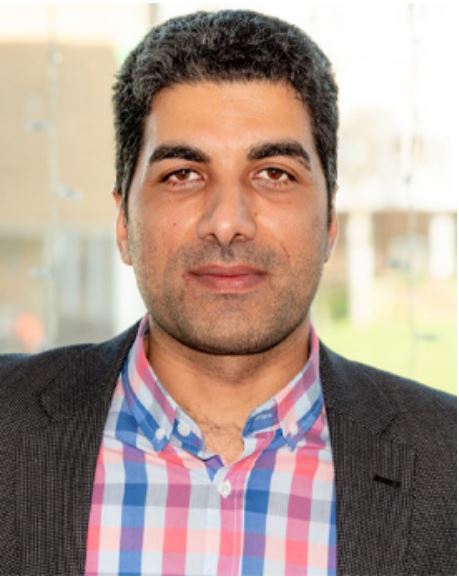}}]{Hossein Rahmani}
received the Ph.D. degree from The University of Western Australia, Perth, WA, Australia, in 2016. He is a Professor with the School of Computing and Communications at Lancaster University in the UK. Before that, he was a Research Fellow with the School of Computer Science and Software Engineering, The University of Western Australia. His research interests include computer vision, action recognition, pose estimation, and deep learning.
\par He now serves as an Associate Editor for the IEEE \textsc{Transactions On Neural Networks And Learning Systems} (TNNLS) and the \textit{Pattern Recognition} (PR) journal, an Area Chair for CVPR 2025 and ICLR 2025. Before that, he served as an Area Chair for CVPR 2024 and ECCV 2024, and a Senior Program Committee member of IJCAI 2023 and 2024.
\end{IEEEbiography}

\begin{IEEEbiography}[{\includegraphics[width=1in,height=1.25in,clip,keepaspectratio]{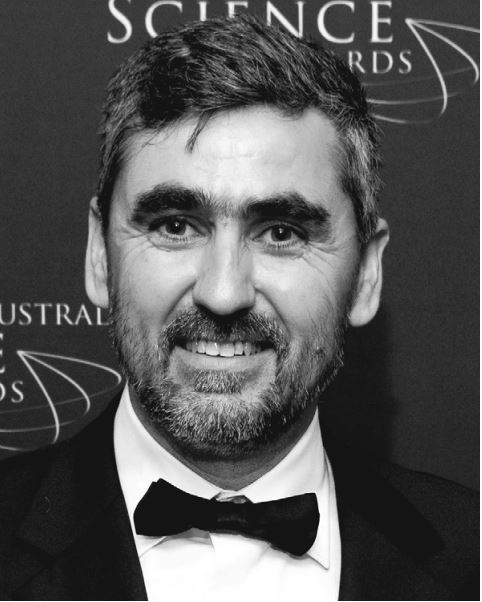}}]{Ajmal Mian (Senior Member, IEEE)} is currently a Professor of Computer Science at The University of Western Australia, Perth, WA, Australia. His research interests include computer vision, machine learning, and artificial intelligence. He has received several awards, including the Vice-Chancellors Mid-Career Research Award and the IAPR Best Scientific Paper Award, etc. He has received several major research grants from the Australian Research Council (ARC) and the U.S. Department of Defense.
\par Prof. Mian is an ARC Future Fellow, an IAPR Fellow, an ACM Distinguished Speaker, and the President of the Australian Pattern Recognition Society. He served as the General Chair for ACCV 2018 and DICTA 2024, and also served as the Area Chair for ACM MM 2020, CVPR 2022, ECCV 2022, and ECCV 2024. He is a Senior Editor for the IEEE \textsc{Transactions On Neural Networks And Learning Systems} (TNNLS) and an Associate Editor for the IEEE \textsc{Transactions On Image Processing} (TIP), and the \textit{Pattern Recognition} (PR) journal.
\end{IEEEbiography}


\end{document}